\documentclass[10pt,twoside]{article}
\usepackage{amsmath}
\usepackage[utf8]{inputenc}
\usepackage{url}
\usepackage{dsfont}
\usepackage{graphicx}
\usepackage{yhmath}
\usepackage[table]{xcolor}
\usepackage{mathrsfs} 
\usepackage{amssymb}
\usepackage{url}
\usepackage{makecell}
\usepackage{arydshln}
\usepackage{multirow}
\usepackage{arydshln}
\usepackage{mathtools}
\usepackage{stmaryrd}  
 \usepackage{amsthm,amssymb}
 \usepackage{subcaption}
 \usepackage{lipsum}
 \usepackage{nccmath}

\input epsf
\def\limiten{\renewcommand{\arraystretch}{0.5}
\begin{array}[t]{c}\stackrel{}{\longrightarrow} \\
{\scriptstyle n\rightarrow
\infty}\end{array}\renewcommand{\arraystretch}{1}}

\numberwithin{equation}{section}

\newtheorem{thm}{Theorem}[section]

\newtheorem{Def}[thm]{Definition}

\newtheorem{rmrk}[thm]{Remark}

\newcommand{\E}{\ensuremath{\mathbb{E}}}
\newcommand{\R}{\ensuremath{\mathbb{R}}}
\newcommand{\Z}{\ensuremath{\mathbb{Z}}}

\newcommand{\N}{\ensuremath{\mathbb{N}}}

\newcommand{\cov}{\ensuremath{\mathrm{Cov}}}
\newcommand{\var}{\ensuremath{\mathrm{Var}}}

\newcommand{\lip}{\ensuremath{\mathrm{Lip}}}
\definecolor{grisclair}{gray}{0.9}
\font\dsrom=dsrom10 scaled 1200
\def \ind{\textrm{\dsrom{1}}}
\DeclareMathOperator*{\argmin}{argmin}

\renewcommand{\arraystretch}{.8}

\begin{document}

\title{\bf Robust deep learning from weakly dependent data}
 \maketitle \vspace{-1.0cm}

\begin{center}
      William Kengne
   and 
     Modou Wade
 \end{center}

  \begin{center}
  { \it 
  Institut Camille Jordan, Université Jean Monnet, 23 Rue Dr Paul Michelon 42023 Saint-Étienne Cedex 2, France\\  
 THEMA, CY Cergy Paris Université, 33 Boulevard du Port, 95011 Cergy-Pontoise Cedex, France\\
  E-mail:   william.kengne@univ-st-etienne.fr  ; modou.wade@cyu.fr\\
  }
\end{center}
 \pagestyle{myheadings}

\markboth{Robust deep learning from weakly dependent data}{Kengne and Wade}

\medskip

\textbf{Abstract}:
Recent developments on deep learning established some theoretical properties of deep neural networks estimators.
 However, most of the existing works on this topic are restricted to bounded loss functions or (sub)-Gaussian or bounded input.
This paper considers robust deep learning from weakly dependent observations, with unbounded loss function and unbounded input/output.
It is only assumed that the output variable has a finite $r$ order moment, with $r >1$.
Non asymptotic bounds for the expected excess risk of the deep neural network estimator are established under strong mixing, and $\psi$-weak dependence assumptions on the observations.
We derive a relationship between these bounds and $r$, and when the data have moments of any order (that is $r=\infty$), the convergence rate is close to some well-known results.
When the target predictor belongs to the class of H\"older smooth functions with sufficiently large smoothness index, the rate of the expected excess risk for exponentially strongly mixing data is close to or as same as those for
obtained with i.i.d. samples.
Application to robust nonparametric regression and robust nonparametric autoregression are considered.
The simulation study for models with heavy-tailed errors shows that, robust estimators with absolute loss and Huber loss function outperform the least squares method. 
 
\medskip
 

\medskip

\section{Introduction}
In the last few years, several significant works have contributed to understand the theoretical properties of deep neural networks (DNN) estimators.
It has been shown that, the excess risk of the DNN predictors obtained from the empirical risk minimization can achieve an optimal convergence rate for regression and classification tasks.
There is a vast literature that focuses on independent and identical distribution (i.i.d.) observations, see, for instance  \cite{ohn2019smooth}, \cite{bauer2019deep}, \cite{schmidt2019deep}, \cite{schmidt2020nonparametric}, 
\cite{tsuji2021estimation}, \cite{ohn2022nonconvex}, \cite{imaizumi2022advantage}, \cite{jiao2023deep}, and the references therein.
We refer to \cite{kengne2024sparse}, \cite{kengne2024deep}, \cite{kengne2023excess}, \cite{kurisu2022adaptive}, \cite{ma2022theoretical}, \cite{kohler2023rate}, \cite{chen2019bbs} for some results on DNNs estimators, obtained with dependent or non-i.i.d. data.
 Despite the significant advances made in deep learning
theory, most results, especially in regression problem, have been obtained with least squares loss function.
Also, the aforesaid works are restricted to bounded loss functions or (sub)-Gaussian or (sub)-exponential or bounded input/output.
Some works in the literature perform robust deep learning, with  theoretical evidence on such predictors; see, among other papers, \cite{shen2021robust}, \cite{lederer2020risk}, \cite{zhang2024generalization}, \cite{wang2022robust}, \cite{kim2023robust}, \cite{xu2023non}.
However, these works focus on i.i.d. observations.
 
\medskip

In this contribution, we consider the observations $ D_n \coloneqq  \{ (X_1, Y_1), \cdots, (X_n, Y_n) \} $ (the training sample) generated from a stationary and ergodic process $\{Z_t =(X_t, Y_t), t \in \Z \} $, which takes values in $\mathcal{Z}=\mathcal{X}  \times \mathcal{Y}$,
where $\mathcal{X} \subset \R^d $ (with $d \in \N$) is the input space and $\mathcal{Y}  \subset \R$ the output space.
A broad class of loss function $\ell: \R \times  \mathcal{Y} \rightarrow [0, \infty) $ is considered in the sequel. 
For a predictor $h \in \mathcal{F} \coloneqq \mathcal{F} (\mathcal{X}, \mathcal{Y})$ (where $\mathcal{F} (\mathcal{X}, \mathcal{Y})$ denotes the set of measurable functions from $ \mathcal{X}  ~ \text{to} ~\mathcal{Y}$), its risk is defined by,
\begin{equation}\label{equation_risk}
 R(h) =  \E_{Z_0} [ \ell (h(X_0), Y_0)] ~ \text{with} ~ Z_0 = (X_0, Y_0).  
\end{equation}
We aim to construct a predictor $h\in \mathcal{F}$ such that, $h(X_t)$ is average "close" to $Y_t$, for all $t\in \Z$, that is, the learner which achieves the smallest risk.
But, the risk defined above (in (\ref{equation_risk})) cannot be minimized in practice, because the distribution of $Z_0 = (X_0, Y_0)$ is unknown.
Since this risk can be effectively estimated by its empirical version, we consider the minimization of the empirical risk, define for all $h \in  \mathcal{F} $ by,
\begin{equation}\label{risk_empirik}
\widehat{R}_n(h) = \dfrac{1}{n}\sum_{i = 1}^n \ell(h(X_i), Y_i).
\end{equation}
A target predictor (assumed to exist) $h^{*}\in \mathcal{F}$, satisfies
\begin{equation}\label{best_pred_F}
R(h^{*}) = \underset{h \in \mathcal{F}}{\inf} R(h).
\end{equation}
The excess risk for any predictor $h \in \mathcal{F}$ is given by,
\begin{equation}
\mathcal{E}_{Z_0}(h) = R(h) - R(h^{*}), ~ \text{with} ~ Z_0 = (X_0, Y_0).
\end{equation} 
The goal is to construct from the observation $D_n$, a DNN predictor $\widehat{h}_n \in \mathcal{H}_{\sigma}(L_n, N_n, B_n, F_n, S_n)$, with a suitable chosen network architecture $(L_n, N_n, B_n, F_n, S_n)$ where $\mathcal{H}_{\sigma}(L_n, N_n, B_n, F_n, S_n)$ is a class of deep neural network predictors defined in Section \ref{Def_DNNs} (see (\ref{DNNs_Constraint})), such that its excess risk is average "close" to zero.
The DNN predictor $\widehat{h}_n$ obtained from the empirical risk minimization (ERM) algorithm over the class $ \mathcal{H}_{\sigma}(L_n, N_n, B_n, F_n, S_n)$ is defined by, 
\begin{equation}\label{DNN_estimator}
 \widehat{h}_n  = \underset{h \in  \mathcal{H}_{\sigma}(L_n, N_n, B_n, F_n, S_n)}{\argmin}\widehat{R}_n(h).    
\end{equation}
In the sequel, we set $\ell(h, z) = \ell(h(x), y)$ for all $z = (x, y) \in \mathcal{Z} = \mathcal{X}\times \mathcal{Y} $ and $h\in \mathcal{F}(\mathcal{X}, \mathcal{Y})$.
We will be interested in establishing a bound of the excess risk $\mathcal{E}_{Z_0}(\widehat{h}_n)$, and studying how fast it tends to zero.

\medskip

 Robustness is known as an important property of an estimator.
 This robustness property includes the ability of a DNN estimator to be resistant to adversarial attacks, such as, a perturbation in the input data.
 Most of the existing results on DNN estimator do not display this robustness property; and are not applicable, for instance, to problems with heavy-tailed data.
The literature on robust deep learning is not vast. 
Robust deep Learning has been performed by \cite{lederer2020risk}, with Lipschitz-continuous
loss functions, in a general class of feedforward neural networks. 
\cite{shen2021robust} consider robust nonparametric estimation with heavy tailed error distributions and establish a non-asymptotic error bounds for a class DNN estimator with the ReLU (rectified linear unit)
activation function.
Robust estimation for the location function from multi-dimensional functional data has been carried out by \cite{wang2022robust}. 
They propose DNN estimators with ReLU activation function and  an absolutely continuous convex loss function.
 \cite{zhang2024generalization} investigate the generalization error of deep convolutional neural networks with the ReLU activation, for robust regression problems within an information-theoretic learning framework. 
We also refer to \cite{kim2023robust}, \cite{xu2023non} for some recent advances on theoretical guarantees of robust DNN.
However, these results are restricted to i.i.d. data. Moreover, theoretical results on robust deep learning from dependent observations are quite scarce. 
 
\medskip

This new work focuses on robust deep learning from a weakly dependent process $\{Z_t = (X_t, Y_t), t\in \Z \} $ (which takes values in $\mathcal{Z} = \mathcal{X}\times \mathcal{Y} \subseteq \R^d \times \R$), based on a training sample $D_n = (X_1, Y_1), \cdots, (X_n, Y_n)$ and a general loss function $\ell : \R \times \mathcal{Y} \rightarrow [0, \infty)$.
The framework considered includes, regression estimation, classification, times series prediction.
Our main contributions are:
\begin{enumerate}
\item For unbounded weakly dependent process $\big( (X_t, Y_t) \big)_{t \in \Z}$,  and unbounded loss function $\ell$, assumed to be Lipschitz continuous,
we establish non asymptotic bounds for the expected excess risk of the DNN estimators, based on a suitable chosen architecture $L_n, N_n, B_n, F_n, S_n$ of the neural networks class.
The output process $(Y_t)_{t \in \Z}$ can be heavy-tailed with a finite $r$-th moment for some $r>1$.
A relationship between the convergence rate of the excess risk  and $r$ is derived, and when the output process has moments of any order (that is $r=\infty$), this rate is close to some well-known results.  
The neural networks class is performed with a broad class of activation functions. 
\begin{itemize}
\item For strong mixing process, the rate of the excess risk is  $\mathcal{O}\Big( \big(\log n^{(\alpha)}  \big)^{3}  \big( n^{(\alpha)} \big)^{-\frac{s}{s + d}(1-\frac{1}{r}) } \Big)$, where $n^{(\alpha)}$ is given in  (\ref{n_alpha}) and $s$ is the smoothness parameter of the H\"older space of the target function (see Section \ref{excess_risk}).
Thus, when the output variable has moments of any order (that is $r=\infty$) and the observations are i.i.d, this rate is $\mathcal{O}\Big(  n^{-\frac{s}{s + d}  } (\log n )^{3}  \Big)$.
\item For $\psi$-weakly dependent observations, the rate derived also depends on $r$ and it is close or "better" than those obtained by \cite{kengne2024deep}, \cite{kengne2023penalized} (by using a penalized regularization procedure), when the output data have moments of any order.
\end{itemize}
\item We carry out nonparametric regression for strong mixing processes with the Huber and the $L_1$ loss functions. 
The model considered takes into account heavy-tailed errors such as: the $t(2)$ (Student's $t$ distribution with $2$ degrees of freedom), the Cauchy$(0,1)$ (Standard Cauchy distribution with location parameter $0$ and scale parameter $1$).
The results established are applicable to such models.
Simulation study is performed for nonlinear autoregressive models with $t(2)$ and $\mathcal{N}(0, 1)$ errors.
Comparison to the least squares method shows that, the robust estimator based on the Huber and the $L_1$ loss functions outperform the estimator based on the $L_2$ loss. 
 This performance of the robust estimator is quite significant  compared to the least squares one, for heavy-tailed models.       
\end{enumerate}

\medskip

The paper is structured as follows.  Section \ref{asump} presents some notations, the weak dependence structure and the class of DNN predictor considered.  The excess risk bounds of the DNN estimators are provided in Section \ref{excess_risk}. 
Application to the robust nonparametric regression is carried out in Section \ref{General_Roust_nonparametric}. We conducted some numerical experiments in Section \ref{numerical_results} and Section \ref{prove} is devoted to the proofs of the main results.

\section{Notations, assumptions and feedforward neural networks}\label{asump}
\subsection{Notations and assumptions}
Let $d \in \N$, $E_1$ and $E_2$ be subsets of separable Banach spaces equipped with norms $\| \cdot\|_{E_1}$ and $\| \cdot\|_{E_2}$ respectively.
Throughout the sequel, we use the following notations. 
\begin{itemize}
\item $\mathcal{F}(E_1, E_2)$ denotes the set of measurable functions from $E_1$ to $E_2$.
\item  For any $h\in \mathcal{F}(E_1, E_2), \epsilon >0$, $B(h, \epsilon) $ is the ball of radius $\epsilon$ of $\mathcal{F} (E_1, E_2) $ centered at $h$, that is,
\[ B (h, \epsilon) = \big\{ f \in \mathcal{F}(E_1, E_2), ~ \| f - h\|_\infty \leq \epsilon \big\},  \]
where $\| \cdot \|_\infty$ denotes the sup-norm defined in (\ref{def_norm_inf}).
\item  For any $\epsilon > 0$, the $\epsilon$-covering number $\mathcal{N}(\mathcal{H}, \epsilon)$  of $\mathcal{H} \subset \mathcal{F}(E_1, E_2) $, represents the minimal 
number of balls of radius $\epsilon$ needed to cover $\mathcal{H}$; that is,
\begin{equation}\label{epsi_covering_number}
 \mathcal{N}(\mathcal{H}, \epsilon) = \inf\Big\{ m \geq 1 ~: \exists h_1, \cdots, h_m \in \mathcal{H} ~ \text{such that} ~ \mathcal{H} \subset \bigcup_{i=1}^m B(h_i, \epsilon) \Big\}.
\end{equation}
\item We set,
\begin{equation}\label{def_norm_inf}
\| h\|_{\infty} = \sup_{x \in E_1} \| h(x) \|_{E_2}, ~ \| h\|_{\infty,U} = \sup_{x \in U} \| h(x) \|_{E_2},
\end{equation}  
 and
\begin{equation}\label{def_Lip}
\lip_\alpha (h) \coloneqq \underset{x_1, x_2 \in E_1, ~ x_1\ne x_2}{\sup} \dfrac{ \|h(x_1) - h(x_2)\|_{E_2}}{\| x_1- x_2 \|^{\alpha}_{E_1}} ~ \text{for any}  ~ \alpha \in [0, 1], 
\end{equation}
 for  any  function $h: E_1 \rightarrow E_2$ and $U \subseteq E_1$.
 \item For any $\mathcal{K} > 0$ and $\alpha \in [0, 1]$, $\Lambda_{\alpha,\mathcal{K}} (E_1, E_2) $ (or simply $\Lambda_{\alpha, \mathcal{K}} (E_1) $ when $E_2 \subseteq \R$) is defined as the set of functions  $h: E_1^u  \rightarrow E_2$ for some $u \in \N$, satisfies  $\|h\|_{\infty} < \infty$ and  $\lip_\alpha(h) \leq \mathcal{K}$.
When $\alpha = 1$, we set  $\lip_1 (h) = \lip(h) $ and $ \Lambda_{1} (E_1) =\Lambda_{1, 1}(E_1, \R)$. 
\item For any $x \in \R^d$, $x^T$ denotes the transpose of $x$.
\item For any $x \in \R$, $\lfloor x \rfloor$ denotes the greatest integer less than or equal to $x$, and $\lceil x \rceil$ the least integer greater than or equal to $x$.  
\end{itemize}
We will consider the following definition in the sequel, see also  \cite{ohn2019smooth}, \cite{ohn2022nonconvex}.
\begin{Def}\label{def_pwl_quad}
Let a function $g: \R \rightarrow \R$.
\begin{enumerate}
\item $g$ is continuous piecewise linear (or "piecewise linear" for notational simplicity) if it is continuous and there exists $K$ ($K\in \N$) break points $a_1,\cdots, a_K \in \R$ with $a_1 \leq a_2\leq\cdots \leq a_K $ such that, for any $k=1,\cdots,K$, $g'(a_k-) \neq g'(a_k+)$ and $g$ is linear on $(-\infty,a_1], [a_1,a_2],\cdots [a_K,\infty)$.
\item $g$ is locally quadratic if there exits an interval $(a,b)$ on which $g$ is three times continuously differentiable with bounded derivatives and there exists $t \in (a,b)$ such that $g'(t) \neq 0$ and $g''(t) \neq 0$.
\end{enumerate} 
\end{Def}

\medskip

We set the following assumptions on the process $\{ Z_t=(X_t, Y_t), t \in \Z \}$ with values in $ \mathcal{Z}=   \mathcal{X} \times \mathcal{Y} \subset \R^d \times \R$, the loss function 
$\ell : \R \times \mathcal{Y} \rightarrow [0, \infty)$, and the activation function $\sigma: \R \rightarrow \R $.
%
%
\begin{itemize}
\item[\textbf{(A1)}]: There exists a constant $C_{\sigma}>0$ such that the activation function $\sigma$ is $C_{\sigma}$-Lipschitzian. That is, $\lip (\sigma) \leq C_{\sigma}$.
Moreover, $\sigma$ is  either piecewise linear or locally quadratic and fixes a non empty interior segment $I \subseteq [0,1]$ (i.e. $\sigma(z) = z$ for all $z \in I$).
\item[\textbf{(A2)}]: There exists $\mathcal{K}_{\ell} > 0$ such that, the loss function $\ell \in \Lambda_{1, \mathcal{K}_{\ell}}(\R \times \mathcal{Y})$. 
\end{itemize}
Assumption \textbf{(A1)} is satisfied by the ReLU (rectified linear units) activation: $\sigma(x) = \max(x, 0)$, and several other activation functions, see \cite{kengne2023excess}.
The $L_1$ and the Huber loss with parameter $\delta >0$ satisfy Assumption \textbf{(A2)} with $\mathcal{K}_{\ell} = 1$ and $\mathcal{K}_{\ell} = \delta$ respectively. 
For all $y \in \R$ and $y' \in \mathcal{Y}$, recall: 
\begin{itemize}
\item The $L_1$ loss function: $\ell(y,y') = |y-y'|$;
\item The Huber loss with parameter $\delta >0$: 
  $ \ell(y,y') =  \left\{
\begin{array}{ll}
      \frac{1}{2}(y-y')^2 &  \text{ if } ~ |y-y'|\leq \delta , \\
 \delta |y-y'| - \frac{1}{2} \delta^2 &   ~ \text{ otherwise}.
\end{array}
\right.
  $
\end{itemize}

 \medskip
 
We consider a separable Banach space $E$. Let us define the weak dependence structure, see \cite{doukhan1999new} and \cite{dedecker2007weak}. 

\begin{Def}\label{Def}
An $E$-valued process $(Z_t)_{t \in \Z}$ is said to be $(\Lambda_1(E), \psi,\epsilon)$-weakly dependent if there exists a function $\psi:[0, \infty)^2 \times \N^2 \to [0, \infty)$ and a sequence $\epsilon=(\epsilon(r))_{r \in \N}$ decreasing to zero at infinity such that for any $g_1, \;g_2 \in \Lambda_1 (E), ~ \text{with} ~ g_1: E^u \rightarrow \R, \;g_2: E^v \rightarrow \R, ~ (u, v \in \N)$ and for any $u$-tuple $(s_1, \cdots, s_u)$ and any $v$-tuple $(t_1, \cdots, t_v) ~ \text{with} ~ s_1 \leq \cdots \leq s_u \leq s_u + r \leq t_1 \leq \cdots \leq t_v$, the following inequality  is fulfilled:
\[ \vert \cov (g_{1}(Z_{s_1}, \cdots, Z_{s_u}),  g_{2}(Z_{t_1}, \cdots, Z_{t_v})) \vert \leq \psi(\lip(g_1),\lip(g_2), u, v) \epsilon(r), \]
where Cov denotes the covariance.
\end{Def}
%
%

Some examples of well-known weak dependence conditions. 
\begin{itemize}
\item $\psi \left(\lip(g_1),\lip(g_2), u, v \right) = v \lip(g_2) $: the $\theta$-weak dependence, then denote $\epsilon(r) = \theta(r) $;
\item $\psi \left(\lip(g_1),\lip(g_2),u,v \right)= u \lip(g_1) + v \lip(g_2)$: the $\eta$-weak dependence, then denote $\epsilon(r) = 
 \eta(r) $;
\item $\psi \left(\lip(g_1),\lip(g_2), u, v \right)= u v \lip(g_1) \cdot \lip(g_2) $: the $\kappa$- weak dependence, then denote $\epsilon(r) = \kappa(r) $;
\item $\psi \left(\lip(g_1), \lip(g_2), u, v \right) = u \lip(g_1) + v \lip(g_2) + u v \lip(g_1) \cdot \lip(g_2) $: the $\lambda$-weak dependence, then denote $\epsilon(r) = \lambda(r) $. 
\end{itemize}
In the sequel, for each of the four choices of $\psi$ above, set respectively,
\begin{equation}\label{Psi}
\Psi (u, v) = 2 v, ~ \Psi (u, v) = u + v, ~ \Psi (u, v) = u v, ~ \Psi(u, v) = (u + v + u v)/2.
\end{equation}

\medskip

Let us set two weak dependence conditions.
\begin{itemize}
\item[\textbf{(A3)}]: Let  $\Psi: [0,  \infty)^2 \times \N^2 \rightarrow [0, \infty)$ be one of the choices in (\ref{Psi}). The process $\left\{ Z_t = (X_t, Y_t), {t \in \Z} \right\}$  is stationary ergodic and   $(\Lambda_1 (\mathcal{Z}), \psi, \epsilon)$-weakly dependent  such that, there exists $L_1, ~ L_2, ~  \mu \ge 0$ satisfying 
\begin{equation}\label{assump_psi_wd}
\sum_{j \ge 0} \left ( j  + 1 \right)^{k} \epsilon(j) \leq L_1 L_2^k  (k!)^{\mu} \quad \text{for all} ~ k \ge 0.
\end{equation}
\end{itemize}

\medskip

\begin{itemize}
\item[\textbf{(A4)}] The process $\{Z_t = (X_t, Y_t ),  t \in \Z \}$ is stationary ergodic, and strong mixing  with the mixing coefficients satisfying for all $j \geq 0$,
\begin{equation}\label{coef_alpha_mixing}
\alpha(j) = \overline{\alpha} \exp(-c j^{\gamma}), ~ \text{ for some } ~ \overline{\alpha} > 0, \gamma > 0, c > 0.
\end{equation}
\item[\textbf{(A5)}] There exists a positive constant $M > 0$ such that $\E[|Y_0|^r] \leq M < \infty$, for some $r > 1$.
\end{itemize}
Many classical autoregressive processes satisfy \textbf{(A3)} and \textbf{(A4)}, see for instance \cite{doukhan2007probability}, \cite{diop2022statistical} and Section \ref{General_Roust_nonparametric} below. 

\subsection{Feedforward neural networks}\label{Def_DNNs}
A neural network function $h$ with network architecture $(L, \mathbf{p})$, where $L$ denotes the number of hidden layers or depth and $\textbf{p} = (p_0, p_1, \cdots, p_{L + 1}) \in \N^{L + 2}$ called width vector is a composition of functions given as follows,
\begin{equation}\label{h_equ1}
h : \R^{p_0} \rightarrow \R^{p_{L+1}}, \;   x\mapsto h(x) = A_{L+1} \circ \sigma_{L} \circ A_{L} \circ \sigma_{L-1} \circ \cdots \circ \sigma_1 \circ A_1 (x),
\end{equation} 
where $ A_k : \R^ {p_ {k - 1}} \rightarrow \R^ {p_k}$ ($k = 1, \cdots,L + 1$) is a linear affine map, defined by $A_k(x) \coloneqq W_k x + \textbf{b}_k$,  for given  $p_ {k- 1}\times p_k$  weight matrix   $ W_k$   and a shift vector $ \textbf{b}_k \in \R^ {p_k} $, $\sigma_k: \R^{p_k} \rightarrow \R^ {p_k}$ is a nonlinear element-wise activation map, defined for all $z=(z_1,\cdots,z_{p_k})$ by $\sigma_k (z) = (\sigma(z_1), \cdots, \sigma(z_{p_k}))^ {T} $ and  $ \sigma: \R \rightarrow \R$ is an activation function.  
Let $\theta(h)$ denote the vector of parameters of a DNN of the form (\ref{h_equ1}), that is,
\begin{equation} \label{def_theta_h}
\theta(h) \coloneqq \left(vec(W_1)^ {T}, \textbf{b}^{T}_{1}, \cdots,  vec(W_{L + 1})^{T} , \textbf{b}^ {T}_{L+1} \right)^{T}, 
 \end{equation} 
 where $ vec(W)$ denotes the vector obtained by concatenating the column vectors of the matrix $W$. 
 In our setting here, $p_0 = d$ and $p_ {L + 1} = 1$.
 In the sequel, we deal with an activation function $ \sigma: \R \rightarrow \R$ and denote by $\mathcal{H}_{\sigma, d, 1} $, the set of DNNs predictors with $d$-dimensional input and $1$-dimensional output.
For a DNN $h$ of the form (\ref{h_equ1}), set depth($h$)$= L$, and width($h$) = $\underset{1\leq j \leq L} {\max} p_j $ which represents the maximum width (the number of neurons) of hidden layers. For any constants $L, N, B, F  > 0$, set
\[ \mathcal{H}_{\sigma}(L, N, B) \coloneqq \big \{h\in \mathcal{H}_{\sigma, q, 1}: \text{depth}(h)\leq L, \text{width}(h)\leq N, \|\theta(h)\|_{\infty} \leq B \big\},  \]
 and
\begin{equation}\label{DNNs_no_Constraint}
\mathcal{H}_{\sigma}(L, N, B, F) \coloneqq \big\{ h: h\in H_{\sigma}(L, N, B), \| h \|_{\infty, \mathcal{X}} \leq F \big\}. 
\end{equation}
The class of sparsity constrained DNNs with sparsity level $S > 0$ is defined by, 
\begin{equation}\label{DNNs_Constraint}
\mathcal{H}_{\sigma}(L, N, B, F, S) \coloneqq \left\{h\in \mathcal{H}_{\sigma}(L, N, B, F) \; :  \; \| \theta(h) \|_0 \leq S 
  \right\},
\end{equation}
where $\| x \|_0 = \sum_{i=1}^p \ind(x_i \neq 0), ~ \| x\| = \underset{1 \leq i  \leq p}{\max} |x_i |$ for all $x=(x_1, \ldots, x_p)^T \in \R^p$ ($p \in \N$).
Let us note that, the network architecture parameters $L, N, B, F, S$ can depend on the sample size $n$, see Section \ref{excess_risk} below.

\section{Excess risk bound for the DNN estimator }\label{excess_risk}
This section aims to establish an upper bound for the excess risk of the DNN estimator for learning from the strong mixing and the $\psi$-weakly dependent processes.
We provide the convergence rate of the excess when the target predictor $h^{*}$ defined at (see (\ref{best_pred_F})) belongs to the set of H\"older functions. 

\medskip

%
Let $s >0$ and $U \subset \R^d$.
Recall that, the H\"older space $\mathcal{C}^s(U)$ is a class of functions $h : U \rightarrow \R$ such that, for any $\alpha\in \N^d$ with $|\alpha| \leq [s], \|\partial^{\alpha}h\|_{\infty} < \infty$, and for any $\alpha\in \N^d$ with $|\alpha| = [s], \lip_{s - [s]}(\partial^{\alpha}h) < \infty$, ($[x]$ denotes the integer part of $x \in \R$), where
\[\alpha = (\alpha_1, \cdots, \alpha_d), ~ |\alpha| = \sum_{i = 1}^d\alpha_i ~ \text{and} ~ \partial^{\alpha} = \dfrac{\partial^{|\alpha|}}{\partial^{\alpha_1}x_1, \cdots, \partial^{\alpha_d}x_d}. 
\]
This space is equipped with the norm: 
\[\|h\|_{\mathcal{C}^s(U)} = \underset{0\leq \alpha \leq [s]}{\sum}\|\partial^{\alpha}(h)\|_{\infty} + \underset{|\alpha| = [s]}{\sum}\lip_{s - [s]}(\partial^{\alpha}h).
\]
For any $s > 0, U \subset \R^d$ and $\mathcal{K} > 0$, set 
\[\mathcal{C}^{s, \mathcal{K}}(U) = \{h\in \mathcal{C}^s(U), \|h\|_{\mathcal{C}^s(U)} \leq \mathcal{K} \}.
\]

\begin{thm}\label{thm1}
Assume that (\textbf{A1}), (\textbf{A2}), \textbf{(A4)}, \textbf{(A5)}  hold and that  $h^{*} \in \mathcal{C}^{s, \mathcal{K}}(\mathcal{X})$ for some $s, \mathcal{K} > 0$, where $h^{*}$ is defined in (\ref{best_pred_F}). Set 
$L_n = \big(1 - \frac{1}{r} \big) \dfrac{s L_0}{s + d} \log\big(n^{(\alpha)} \big)$, $N_n = N_0 \big(n^{(\alpha)} \big)^{\big(1 - \frac{1}{r} \big) \frac{d}{s + d}}$, $S_n = \big(1 - \frac{1}{r} \big) \frac{s S_0}{s + d} (n^{(\alpha)})^{\big(1 - \frac{1}{r} \big) \frac{d}{s + d}}\log\big(n^{(\alpha)} \big)$,
$ B_n = B_0 \big(n^{(\alpha)} \big)^{ \big(1 - \frac{1}{r} \big) \frac{4s(d/s + 1)}{s + d}}$ and $F_n > \mathcal{K}$, 
for some positive constants $L_0, N_0, S_0, B_0 > 0$. Consider the class of DNN $\mathcal{H}_{\sigma}(L_n, N_n, B_n, F_n, S_n)$ defined in (\ref{DNNs_Constraint}).
Then, there exists $n_0 = n_0(\mathcal{K}_{\ell}, L_0, N_0, B_0, S_0, C_\sigma,$ $ s, d, r, c, \gamma) \in \N$ defined at (\ref{equa_limite}), such that for all $n \geq n_0$, the DNN estimator $\widehat{h}_n$ defined in  (\ref{DNN_estimator}) satisfies
\begin{equation}\label{thm1_excess_risk_bound}
\E[R(\widehat{h}_n) - R(h^{*})] \leq  \dfrac{ \big(\log n^{(\alpha)}  \big)^{\nu} + \mathcal{K}_{\ell} }{ \big( n^{(\alpha)} \big)^{\frac{s}{s + d}(1-\frac{1}{r}) } }   + \frac{ C(\mathcal{K}_{\ell}, \overline{\alpha}, M)  }{ \big(n^{(\alpha)} \big)^{(1 - \frac{1}{r})} }   + \frac{3\mathcal{K}_{\ell}}{n^{(\alpha)}}, 
\end{equation}
for all $\nu>3$, with, 
\begin{equation}\label{n_alpha}
n^{(\alpha)} = \lfloor n \lceil \{8n/c\}^{1/(\gamma + 1)}\rceil^{-1} \rfloor ~ \text{ and }  C(\mathcal{K}_{\ell}, \overline{\alpha}, M) =   \frac{64}{3}\mathcal{K}_{\ell}  (1 + 4e^{-2}\overline{\alpha})  +  6\mathcal{K}_{\ell} M,
\end{equation}
where $c, \gamma, \overline{\alpha}$ are given in (\ref{coef_alpha_mixing}), $\mathcal{K}_{\ell}$ in the assumption (\textbf{A2}) and $r, M$ in the assumption (\textbf{A5}).
\end{thm}
\noindent
Therefore, the rate of the expected excess risk in the bound (\ref{thm1_excess_risk_bound}) is $\mathcal{O}\Big( \big(\log n^{(\alpha)}  \big)^{3}  \big( n^{(\alpha)} \big)^{-\frac{s}{s + d}(1-\frac{1}{r}) } \Big)$.

\begin{rmrk} \hfill
\begin{enumerate}
\item  For $n \ge \max(c/8, 2^{2 + 5/\gamma}c^{1/\gamma})$, we have $n^{(\alpha)} \ge C(\gamma, c)n^{\gamma/(\gamma+1)}$, with $C(\gamma,c) = 2^{-\frac{2\gamma + 5}{\gamma + 1}}c^{\frac{1}{\gamma + 1}}$, see also \cite{hang2014fast}.
So, the rate in (\ref{thm1_excess_risk_bound}) is  $\mathcal{O}\Big(  n^{-\frac{s}{s + d}(1-\frac{1}{r}) \frac{\gamma}{\gamma + 1} } (\log n )^{3}  \Big)$.
\item Therefore, for of i.i.d data (that is $\gamma = \infty$),  this rate is close (up to logarithmic factor) to that obtained in some specific case in \cite{shen2021robust} (see Corollary 2).
\item If the output variable has moments of any order (that is $r=\infty$) and the observations are i.i.d, then, the rate above is $\mathcal{O}\Big(  n^{-\frac{s}{s + d}  } (\log n )^{3}  \Big)$. 
Recall that, the rate obtained in the nonparametric regression with square loss is $\mathcal{O}\Big(  n^{-\frac{2s}{2s + d}  } (\log n )^{3}  \Big)$, see for instance \cite{schmidt2020nonparametric}. 
\end{enumerate}
\end{rmrk}

\medskip

\noindent In the sequel, we consider an architecture $L_n, N_n, B_n, F_n, S_n$ as in Theorem \ref{thm1} and we let $\beta_n \ge F_n$.
Set $C_{n, 1} = 32\mathcal{K}_{\ell}^2 \beta_n^2\Psi(1, 1)L_1$ and $C_{n, 2} = 8\mathcal{K}_{\ell} \beta_nL_2\max(\dfrac{2^{3 + \mu}}{\Psi(1, 1)}, 1)$, where $\Psi$ is one of the functions defined at (see (\ref{Psi})). 
\begin{thm}\label{thm2}
Assume (\textbf{A1})-(\textbf{A3}), \textbf{(A5)} and $h^{*} \in \mathcal{C}^{s, \mathcal{K}}(\mathcal{X})$ for some $ \mathcal{K} > 0$ and $s > \max\Big\{ d \big( \frac{(r-1)(2\mu+3)(\mu +2)}{ r(2\mu +3) - (\mu+1)} - 1 \big),  d \big( \big(1 - \frac{1}{r} \big)(2\mu +3) -1 \big) \Big\}$. 
 Let 
$L_n = \big(1 - \frac{1}{r} \big) \dfrac{s L_0}{s + d} \log(n ), N_n = N_0 n^{\big(1 - \frac{1}{r} \big) \frac{d}{s + d}}, S_n = \big(1 - \frac{1}{r} \big) \frac{s S_0}{s + d} n^{\big(1 - \frac{1}{r} \big) \frac{d}{s + d}}\log(n),  B_n = B_0 n^{ \big(1 - \frac{1}{r} \big) \frac{4s(d/s + 1)}{s + d}},
$
and $F_n \leq n^{\frac{\mu + 1}{r(2\mu + 3)} }$,
for some $L_0, N_0, B_0, S_0 >0$. 
Consider the class of DNN $\mathcal{H}_{\sigma}(L_n, N_n, B_n, F_n, S_n)$ defined in (\ref{DNNs_Constraint}).
Then, there exists $n_0 = n_0(\mathcal{K}_{\ell}, L_0, N_0, B_0, S_0, C_\sigma,$ $ s, d, r, \mu) \in \N$, such that for all $n \geq n_0$, the DNN estimator $\widehat{h}_n$ defined in  (\ref{DNN_estimator}) satisfies,
\begin{equation}\label{thm2_bound}
\E[R(\widehat{h}_n) - R(h^{*})] 
 \leq  \dfrac{C(\mu, \mathcal{K}_{\ell}, L_1, L_2, M)}{n^{\big(1 - 1/r \big)(\mu + 1)/(2\mu + 3)}}  + \dfrac{3\mathcal{K}_{\ell}}{n} + \dfrac{ 2\mathcal{K}_{\ell}}{ n^{\frac{s}{s + d}(1-1/r) } } , 
\end{equation}
for some  positive constant $C(\mu, \mathcal{K}_{\ell}, L_1, L_2, M)$ depending on $\mu, \mathcal{K}_{\ell}, L_1, L_2, M$, where $\mu, L_1, L_2$ are given in the assumption (\textbf{A3}), $\mathcal{K}_{\ell}$ in \textbf{(A2)} and $M$ in \textbf{(A5)}. 
\end{thm}
\begin{rmrk}
For some classical models such as AR($p$), ARCH($p$), assumption (\textbf{A3}) holds with $\mu =0$.
In this case, if $s \gg d$, then the convergence rate in (\ref{thm2_bound}) is close to $\mathcal{O}\big( n^{-(1 - 1/r)/3} \big)$.
If the output variable has moments of any order (that is $r=\infty$), this rate is close to $\mathcal{O}(n^{-1/3})$.
\end{rmrk}
\section{Robust nonparametric regression}\label{General_Roust_nonparametric}
In this section we deal with the robust nonparametric regression, where the output $Y_t \in \R$ and the input $X_t \in \mathcal{X} \subseteq \R^d$ generated from the model:
\begin{equation}\label{robust_regression}
Y_t = h^{*}(X_t) + \epsilon_t.
\end{equation}
Where $h^{*} :\mathcal{X}\rightarrow \R$ is an unknown regression function, $\epsilon_t$ is an error variable independent to $X_t$. We assume that the error has a finite $r$-th moment for some $r > 1$, that is $\E|\epsilon_t|^r < \infty$.  
Let us check if the model (\ref{robust_regression}) satisfies the assumptions of Theorem \ref{thm1}. 
\begin{itemize}
\item[$\bullet$] We deal with the ReLU activation function, thus the assumption \textbf{(A1)} holds.
\item[$\bullet$]  We consider the $L_1$ and the Huber (with parameter $\delta = 1.345$) loss functions, so, the assumption \textbf{(A2)} is satisfied.
\item[$\bullet$] It is assumed that $h^{*} :\mathcal{X}\rightarrow \R$ is a smooth function, that is, $h^{*} \in \mathcal{C}^{s, \mathcal{K}}(\mathcal{X})$ for some $s, \mathcal{K} > 0$. 
\item[$\bullet$] We assume that the process $Z_t = (X_t, Y_t)_{t\in \Z}$ is  $(\Lambda_1 (\mathcal{Z}), \psi, \epsilon)$-weakly dependent where $\epsilon=\big(\epsilon(j)\big)_{j \in \N}$ is such that (\ref{assump_psi_wd}) holds for some $L_1, ~ L_2, ~  \mu \ge 0$ or is strong mixing 
with $\alpha$-mixing coefficients defined by
$\alpha_{Z}(j) = \overline{\alpha}\exp(-cj^{\gamma}), j \ge 1, \overline{\alpha} > 0, \gamma > 0, c > 0$. That is, \textbf{(A3)} or \textbf{(A4)} holds.   
\item[$\bullet$] If $\mathcal{X}$ is a compact set of $\R^d$, then, one can easily see that \textbf{(A5)} is verified when (i) $\epsilon_t \sim \mathcal{N}(0, 1)$ denoted the standard normal distribution, $\E|\epsilon_t|^r < \infty, r\in [0, \infty[$; 
(ii) $\epsilon_t \sim t(2)$ denoted the Student's $t$ distribution with 2 degrees of freedom, $\E|\epsilon_t|^r < \infty, r\in [0, 2)$; (iii) $\epsilon_t \sim \text{Cauchy}(0, 1)$ denoted the Standard Cauchy distribution with location parameter $0$ and scale parameter $1$, $\E|\epsilon_t|^r < \infty, r\in [0, 1]$. 
Hence, the result of the Theorem \ref{thm1} can be applied to the model (\ref{robust_regression}).
\end{itemize}
\subsection*{Application to autoregressive models}
We consider the nonlinear autoregressive model 
\begin{equation}\label{nonlinear_autoregressive}
Y_t = f(Y_{t - 1}, \cdots, Y_{t - p}) + \epsilon_t,  
\end{equation}
for some measurable unknown regression function  $f : \R^{p}\rightarrow \R ~ (p\in \N)$ and $(\epsilon_t)_{t\in \Z}$ is an i.i.d. process.
Set $X_t = (Y_{t - 1}, \cdots, Y_{t - p})$, one can see that the model $(\ref{nonlinear_autoregressive})$ is a particular case of the model (\ref{robust_regression}). 
%
%
%
%
Let us check assumptions (\textbf{A3}) or (\textbf{A4}), and (\textbf{A5}).
For this purpose, consider the following assumptions: 
\begin{itemize}
\item[A($\epsilon$):] The sequence $(\epsilon_t)$ has positive and continuous density function everywhere, $\E[\epsilon_t] = 0$, and $\epsilon_t$ is independent of $X_{t - s}, ~ s > 0$;
\item[A$_1(f)$:] The function $f$ is bounded on every compact of $\R^p$, that is, for all $K \ge 0$, 
\[ \underset{\|x\| \leq K}{\sup} |f(x)| < \infty, 
\]
where $\|\cdot\|$ denotes the Euclidean norm on $\R^p$;
\item[A$_2(f)$:] There exist constants $\alpha_i \ge 0, ~ i = 1, \cdots, p, M > 0, c_1 > 0$ such that 
\begin{equation}
|f(x)| \leq \sum_{i = 1}^p \alpha_i |x_i| + c_1, ~ \|x\| \ge M.
\end{equation}.
\end{itemize}
Under the assumptions A($\epsilon$), A$_1(f)$, A$_2(f)$, if the following condition holds
\begin{equation}\label{auto_reg_cond_alpha_i}
\sum_{i = 1}^p \alpha_i < 1.
\end{equation}
One can easily see in \cite{chen2000geometric}, \cite{doukhan1994mixing}, \cite{chen2018generalized}, there exists a stationary solution $(Y_t, X_t)_{t\in \Z}$ of the model (\ref{nonlinear_autoregressive}) which is geometrically strong mixing. Thus, the assumption (\textbf{A4}) holds.
Also, under the condition (\ref{auto_reg_cond_alpha_i}), the process $(Y_t, X_t)_{t\in \Z}$ solution of (\ref{nonlinear_autoregressive}), is $\theta$-weakly dependent such that (\textbf{A3}) holds with $\mu =2$, see  \cite{diop2022statistical} and \cite{kengne2024deep}.    
Hence, the results of the Theorem \ref{thm1} and Theorem \ref{thm2} can be applied to the model (\ref{nonlinear_autoregressive}). 

\section{Numerical results}\label{numerical_results}
In this section, we carry out some numerical experiments to assess the performance of  DNNs predictors for the estimation of nonlinear autoregressive time series.

\medskip

Let $(Y_1, X_1), \cdots, (Y_n, X_n)$ be a trajectory of the process $\left\{ (X_t, Y_t), {t \in \Z} \right\}$ in (\ref{nonlinear_autoregressive}).
We will focus to the prediction of $Y_{n + 1}$ from this training sample. We perform the learning theory with DNNs estimators develop above, with the input variable 
$ X_t = (Y_{t - 1},\cdots, Y_{t - p})$, the input space $\mathcal{X} \subset \R^p$, and the output space $ \mathcal{Y} \subset \R$, where $p=3, 2$ in DGP1 and DGP2 respectively (see below). The following models are specific cases of (\ref{nonlinear_autoregressive}): 
\begin{equation*}
\begin{array}{ll}
\text{DGP1} : f(Y_{t - 1}, Y_{t - 2}, Y_{t - 3}) & = 0.5 - 0.5\max(Y_{t - 1}, 0) + 0.2\min(Y_{t - 1}, 0)  + 0.15Y_{t - 3} + \epsilon_t,
\\
\text{DGP2} : f(Y_{t - 1}, Y_{t - 2}) & = 0.75 + \Big(0.8 - 0.2e^{-Y_{t - 1}^2} \Big) Y_{t - 1} +  \Big(-0.2 + 0.3e^{-Y_{t - 1}^2} \Big) Y_{t - 2} + \epsilon_t, 
\end{array}
\end{equation*}
where  $(\epsilon_t)_{t\in \Z}$ represents an innovation generated respectively from the Student's distribution with 2 degrees of freedom, denoted by $\epsilon_t \sim t(2)$ and the standard normal distribution denoted by $\epsilon_t \sim\mathcal{N}(0, 1)$ for each DGPs, with $s = 1, 2$. 
DGP1 is  a threshold autoregressive model, whereas DGP2 is an exponential autoregression model.

\medskip

We deal with the $L_2$, Huber and the $L_1$ loss functions.
The target function with respect to the $L_2$ loss is given for all $x \in \mathcal{X}$ by:
\begin{equation}\label{target_L2}
h^*(x) = \E[Y_0| X_0 = x] = f(x).   
\end{equation}
Since the distribution of $(\epsilon_t)_{t\in \Z}$ is symmetric around $0$, the target predictor with respect to the Huber and $L_1$ loss functions (see also \cite{fan2022noise}, \cite{shen2021robust}) are given respectively for all $x \in \mathcal{X}$ by: 
\begin{equation}\label{target_L2}
h^*(x) = f(x),   
\end{equation}
and,
\begin{equation}\label{target_L1}
h^*(x) = \text{med}(Y_0| X_0 = x) = f(x),   
\end{equation}
 where med denotes the median.

\medskip

For each of these DGPs, we perform a network architecture of 2 hidden layers with 100 hidden nodes for each layer. The ReLU and the linear activation function are respectively used in the hidden layer and the output layer. The optimization algorithm Adam (see cite (\cite{kingma2014adam})) with learning rate $10^{-3}$ is used to train the network's weights and the minibatch size of 32. 
For each loss, we stopped the training when the empirical error of the DNN estimator is not improved over 30 epochs.

\begin{figure}[h!]
  \vspace{-1cm} 
 \begin{center}
  \includegraphics[height=22.1cm, width=17.5cm]{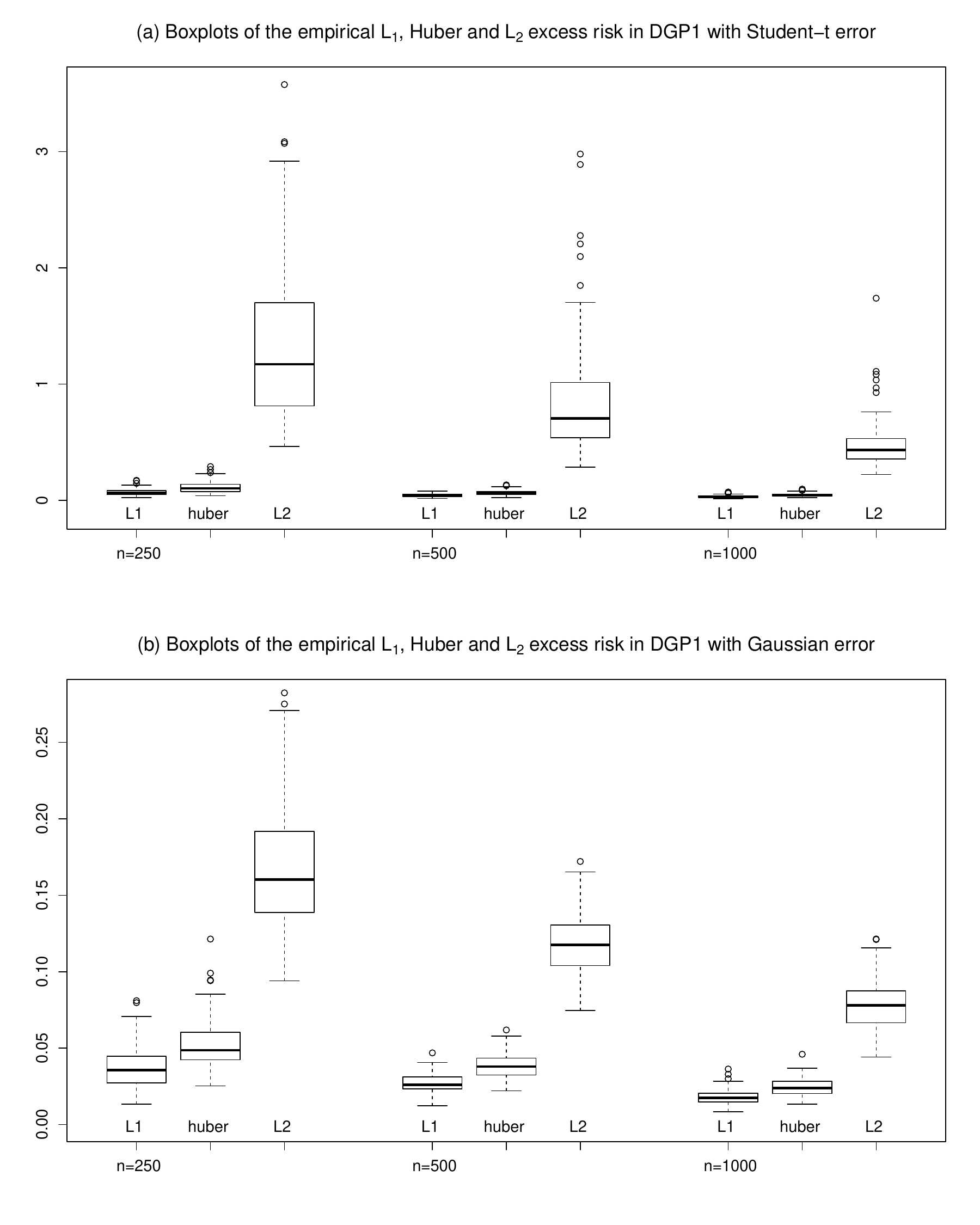} 
  \end{center}  
\vspace{-1.2cm}
\caption{\it Boxplots of the empirical $L_1$, Huber and $L_2$ excess risk of the DNN predictors with $n=250, 500$ and 1000 in DGP1 with Student-t error (a) and Gaussian error (b).}
\label{Graph_excess_risk_DGP1}
\end{figure}

\begin{figure}[h!]
  \vspace{-1cm} 
 \begin{center}
  \includegraphics[height=21.9cm, width=17.5cm]{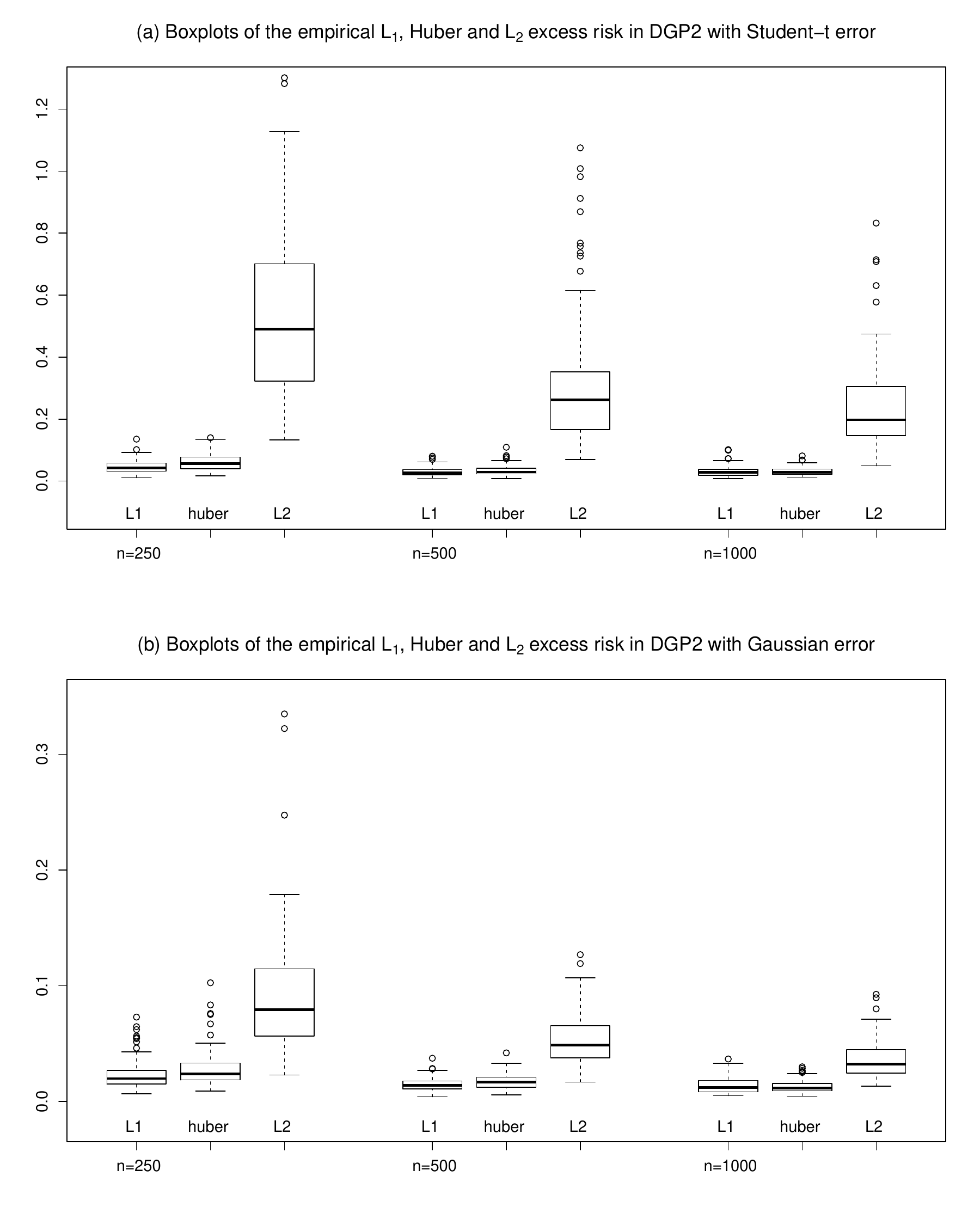} 
  \end{center}  
\vspace{-1.2cm}
\caption{\it Boxplots of the empirical $L_1$, Huber and $L_2$ excess risk of the DNN predictors with $n=250, 500$ and 1000 in DGP2 with Student-t error (a) and Gaussian error (b).}
\label{Graph_excess_risk_DGP2}
\end{figure}

\medskip
For $n = 250, 500$, and $1000$, we generated a trajectory $((Y_1, X_1), \cdots, (Y_n, X_n))$ from the true DGP and the predictor $\widehat{h}_n$ is obtained from $(\ref{DNN_estimator})$.
 The empirical $L_1$, Huber, $L_2$ excess risk of $\widehat{h}_n$ is evaluated from a new trajectory  $((Y_1', X_1'), \cdots, (Y_m', X_m'))$ with $m = 10^4$.
Figure \ref{Graph_excess_risk_DGP1} and Figure \ref{Graph_excess_risk_DGP2} show the boxplots of the empirical $L_1$, Huber and $L_2$ excess risk of the DNN estimator based on 100 replications. 
For each DGPs ($s = 1, 2$), one can see that, the empirical excess risk decreases when the sample size $n$ increases.
 These numerical findings with the $L_1$ and Huber loss are in accordance with Theorem \ref{thm1} and Theorem \ref{thm2}.

\medskip

\begin{figure}[h!]
  \vspace{-1cm} 
 \begin{center}
  \includegraphics[height=21.85cm, width=17.5cm]{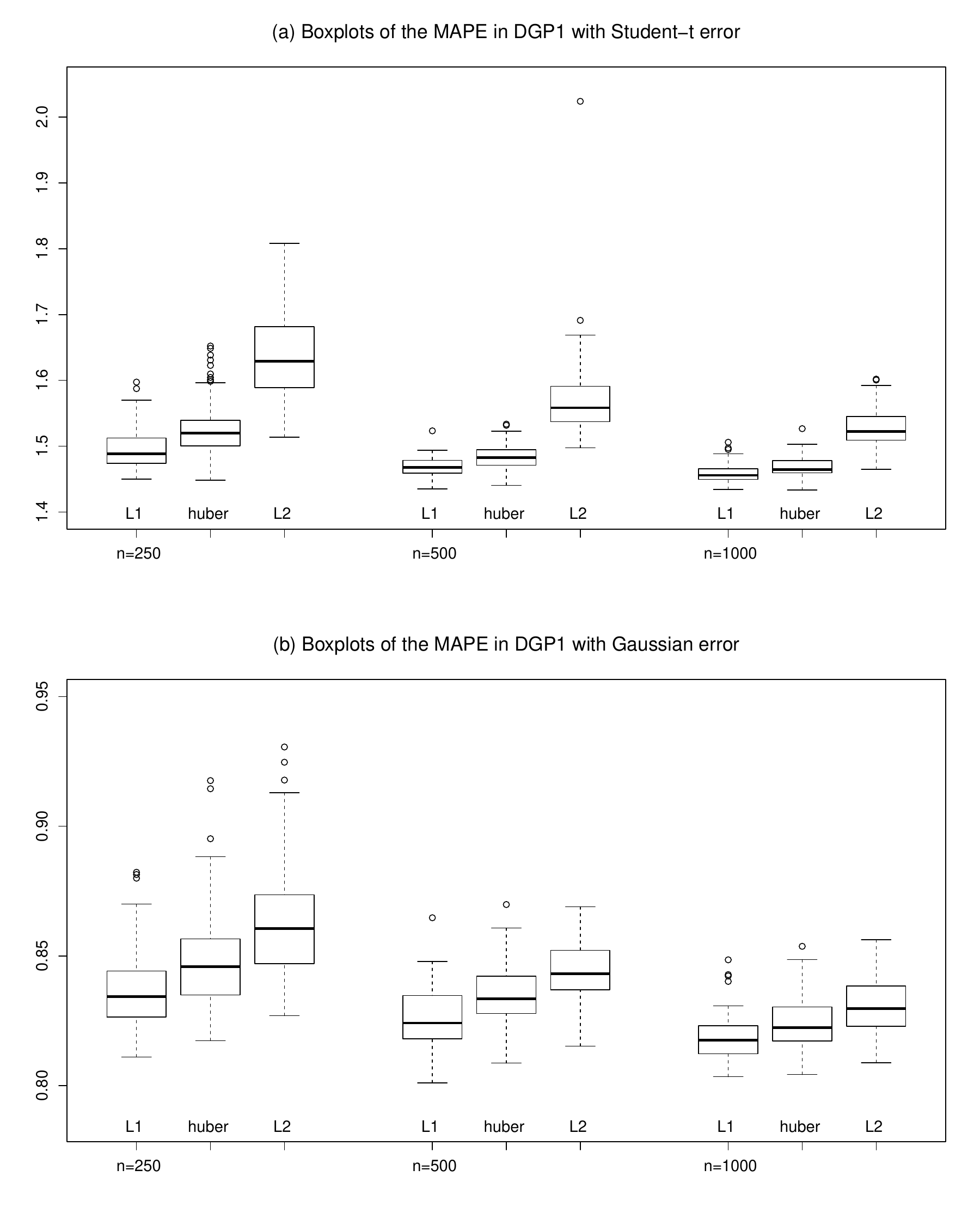} 
  \end{center}  
\vspace{-1.2cm}
\caption{\it Boxplots of the mean absolute prediction error of the DNN predictors with $n=250, 500$ and 1000 in DGP1 with Student-t error (a) and Gaussian error (b).}
\label{Graph_MAPE_DGP1}
\end{figure}

\begin{figure}[h!]
  \vspace{-1cm} 
 \begin{center}
  \includegraphics[height=21.85cm, width=17.5cm]{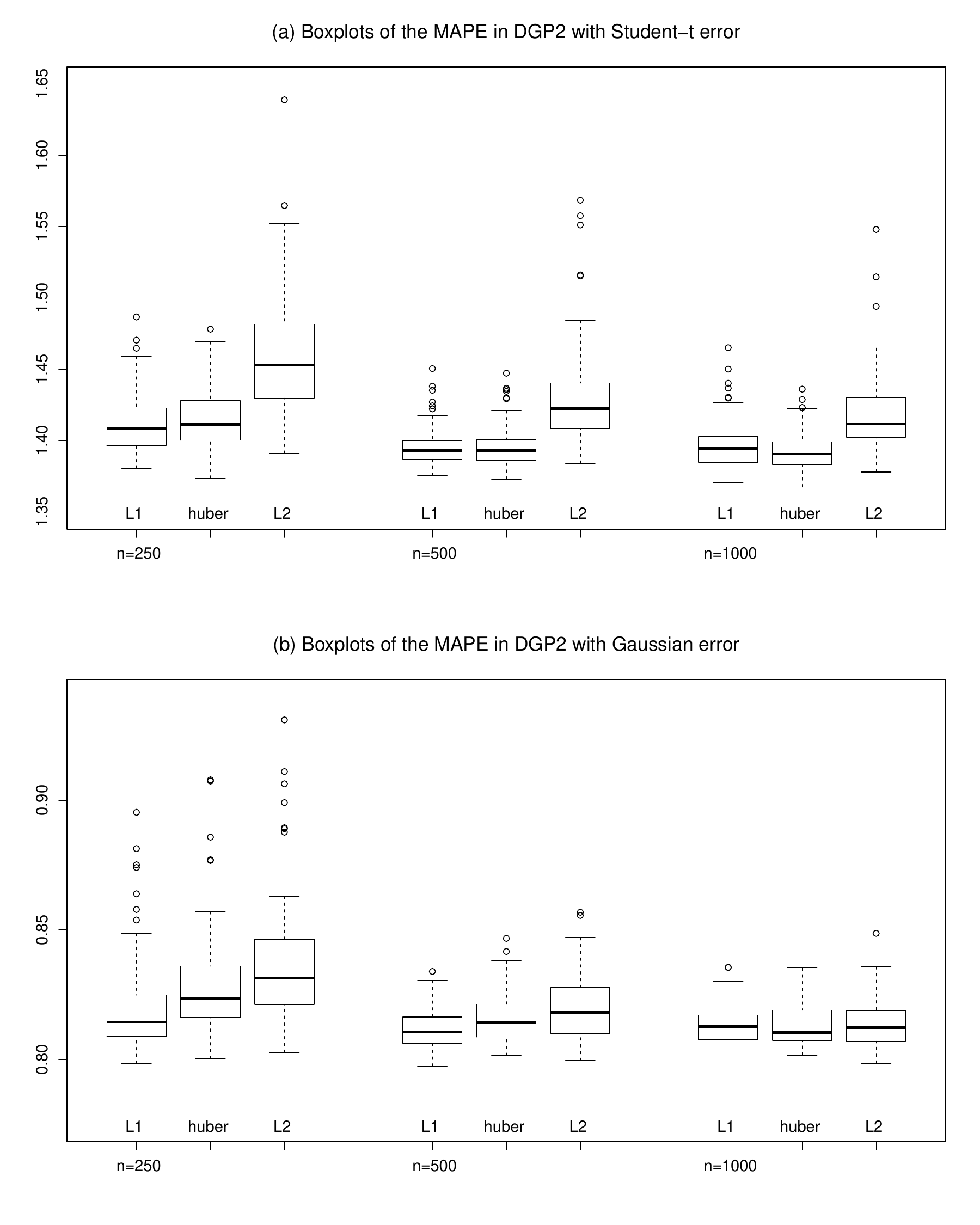} 
  \end{center}  
\vspace{-1.2cm}
\caption{\it Boxplots of the mean absolute prediction error of the DNN predictors with $n=250, 500$ and 1000 in DGP2 with Student-t error (a) and Gaussian error (b).}
\label{Graph_MAPE_DGP2}
\end{figure}

\begin{figure}[h!]
  \vspace{-0.5cm} 
 \begin{center}
  \includegraphics[height=21.35cm, width=17.5cm]{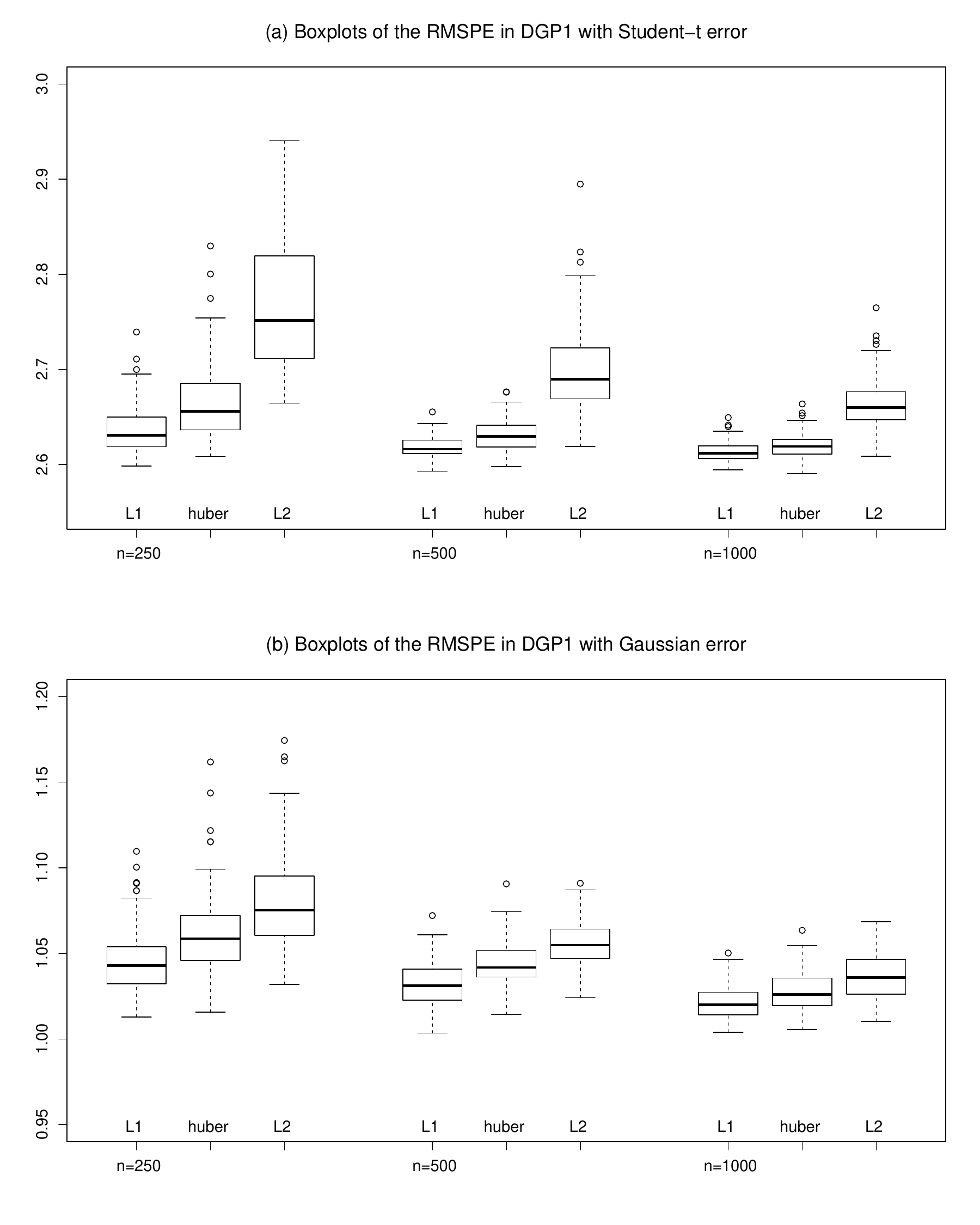} 
  \end{center}  
\vspace{-1.2cm}
\caption{\it Boxplots of the root mean square prediction error of the DNN predictors with $n=250, 500$ and 1000 in DGP1 with Student-t error (a) and Gaussian error (b).}
\label{Graph_RMSPE_DGP1}
\end{figure}

\begin{figure}[h!]
  \vspace{-0.5cm} 
 \begin{center}
  \includegraphics[height=21.35cm, width=17.5cm]{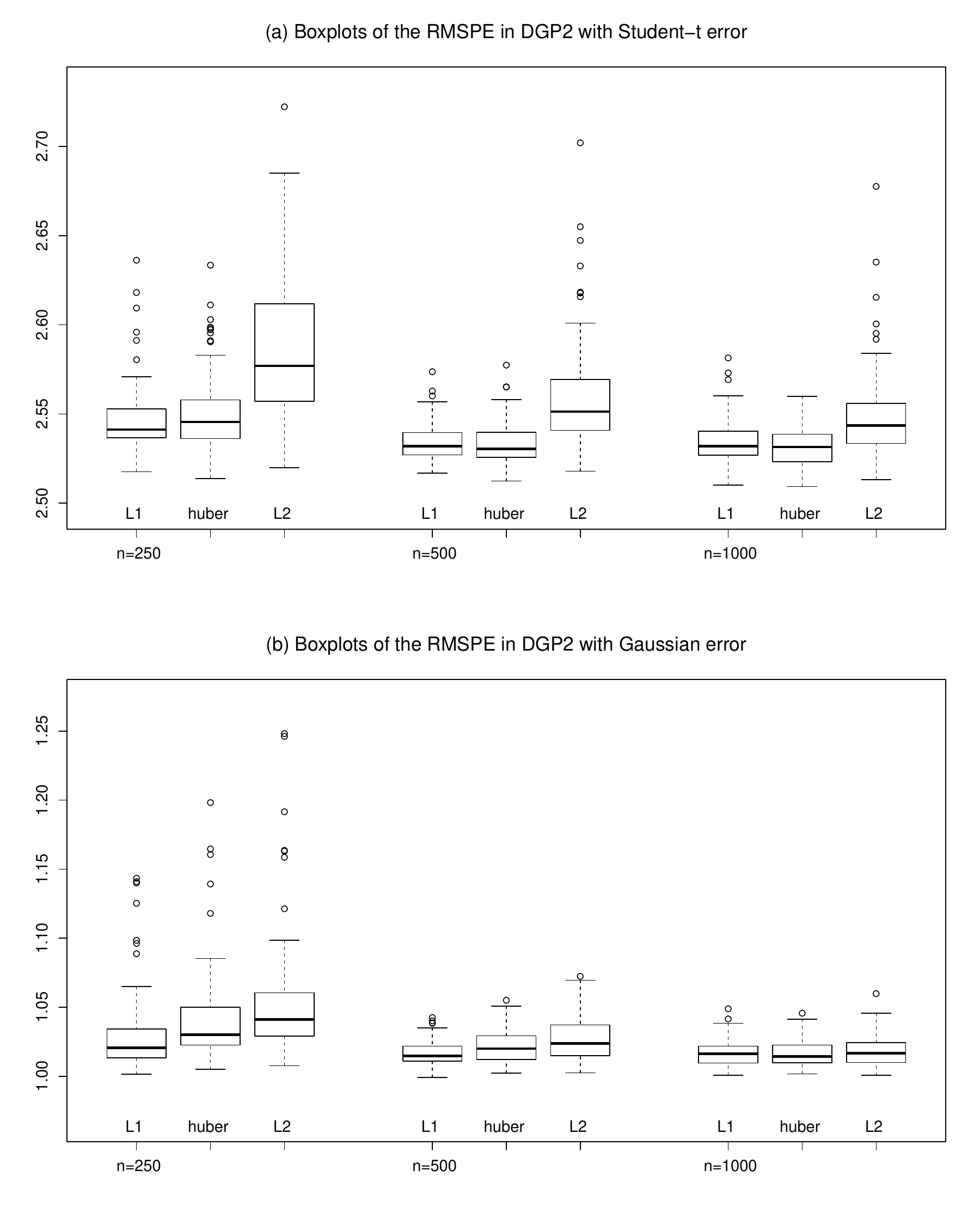} 
  \end{center}  
\vspace{-1.2cm}
\caption{\it Boxplots of the root mean square prediction error of the DNN predictors with $n=250, 500$ and 1000 in DGP2 with Student-t error (a) and Gaussian error (b).}
\label{Graph_RMSPE_DGP2}
\end{figure}

To compare the performance of the estimators based on the $L_1$, Huber and $L_2$ loss functions, we consider the mean absolute prediction error (MAPE) and the root mean square prediction error (RMSPE).
For each training sample $((Y_1, X_1), \cdots, (Y_n, X_n))$, a test sample $((Y''_1, X''_1), \cdots, (Y''_n, X'_n))$ (independent of the training one) is generated from the DGP.
The MAPE and RMSPE are given for each $n=250, 500, 1000$ by:
\[MAPE_n = \dfrac{1}{n - p} \sum_{i=p+1}^n \big| Y''_i - \widehat{h}_n(X''_i)  \big|  ; ~  RMSPE_n= \sqrt{\dfrac{1}{n - p} \sum_{i=p+1}^n\big(Y''_i  - \widehat{h}_n(X''_i) \big)^2 }. \] 
Figures \ref{Graph_MAPE_DGP1}, \ref{Graph_MAPE_DGP2} and Figures \ref{Graph_RMSPE_DGP1}, \ref{Graph_RMSPE_DGP2} display the boxplots of the MAPE and RMSPE of the DNN predictors over 100 replications. 
One can see that, the robust estimators based on the absolute and the Huber loss functions outperform the least squares method in DGP1 and in DGP2 for both Student and Gaussian error.
These performances (in terms of MAPE and RMSPE) of the robust estimator are more significant compared to the least squares one, for models with Student's $t(2)$ error.
%
%

\newpage

\newpage

\section{Proofs of the main results}\label{prove}
 \subsection{Proof of Theorem  \ref{thm1}}
 Let us consider two independent sets of observations, $ D_n = \{(X_i, Y_i)_{i = 1}^n)\}$ and $ D_n' = \{(X_i', Y_i')_{i = 1}^n) \}$ generated, respectively, from a stationary ergodic process
$ Z = \mathcal{X} \times \mathcal{Y}$ and $ Z' = \mathcal{X'} \times \mathcal{Y'}$, where $\mathcal{X}, \mathcal{X'} \subset \R^d$, and $\mathcal{Y}, \mathcal{Y'} \subset \R$.
 For all $i=1,\cdots,n$ and a predictor $h:\mathcal{X} \rightarrow \mathcal{Y}$, we set: 
\begin{equation}\label{proof_def_g}
g(h(X_i), Y_i) := \ell(h(X_i), Y_i) - \ell(h^{*}(X_i), Y_i).
\end{equation} 
Recall that, the empirical risk minimizer $\widehat{h}_n$ depends on the sample $D_n$, and its excess risk is given by $ \E_{D_n'} \Big[\dfrac{1}{n}\sum_{i= 1}^n g(\widehat{h}_n, Z_i') \Big]$ and its expected excess risk is defined by
\begin{equation}\label{Exp_excess_risk_v1}
\E[R(\widehat{h}_n) - R(h^{*})] = \E_{D_n} \Big[\E_{D_n'} \Big[\dfrac{1}{n}\sum_{i= 1}^n g(\widehat{h}_n, Z_i') \Big] \Big].
\end{equation}
Let $L_n, N_n, B_n, F_n, S_n >0$ satisfying the conditions in Theorem \ref{thm1}.
In the sequel, we set:
\begin{equation}\label{proof_def_H_sigma_n}
 \mathcal{H}_{\sigma,n}  := \mathcal{H}_{\sigma}(L_n, N_n, B_n, F_n, S_n).
\end{equation} 
We consider a target neural network (assumed to exist) $h_{\mathcal{H}_{\sigma, n}} \in \mathcal{H}_{\sigma, n}$ defined by
\begin{equation}
h_{\mathcal{H}_{\sigma, n}} = \underset{h \in \mathcal{H}_{\sigma, n}}{\argmin} R(h).
\end{equation}
Let us recall the approximation error of $h^{*}$ given by
\[R(h_{\mathcal{H}_{\sigma, n}}) - R(h^{*}). 
\]
Recall that the approximation error depends only on the functions class $\mathcal{H}_{\sigma, n}$ and the distribution of the data.
The definition of the empirical risk minimizer, implies 
\[ \widehat{R}_n(\widehat{h}_n) \leq \widehat{R}_n(h_{\mathcal{H}_{\sigma, n}}).
\]
That is,
\begin{equation}\label{Compa_ERM_Target_Funct}
\E_{D_n} \Big[\dfrac{1}{n}\sum_{i= 1}^n g(\widehat{h}_n, Z_i) \Big] \leq \E_{D_n} \Big[\dfrac{1}{n}\sum_{i= 1}^n g(h_{\mathcal{H}_{\sigma, n}}, Z_i) \Big]. 
\end{equation}
According to (\ref{Compa_ERM_Target_Funct}) and  (\ref{Exp_excess_risk_v1}), we get 
\begin{align}\label{Exp_excess_risk_v2}
\nonumber \E[R(\widehat{h}_n) - R(h^{*})] & = \E_{D_n} \Big[\dfrac{1}{n}\sum_{i= 1}^n \{ -2g(\widehat{h}_n, Z_i) + 2g(\widehat{h}_n, Z_i) + \E_{D_n'} g(\widehat{h}_n, Z_i') \} \Big]
\\
\nonumber & \leq \E_{D_n} \Big[\dfrac{1}{n}\sum_{i= 1}^n \{ -2g(\widehat{h}_n, Z_i)+ \E_{D_n'} g(\widehat{h}_n, Z_i') \} \Big]  + 2\E_{D_n} \Big[\dfrac{1}{n}\sum_{i= 1}^n g(\widehat{h}_n, Z_i) \Big] 
\\
\nonumber & \leq  \E_{D_n} \Big[\dfrac{1}{n}\sum_{i= 1}^n \{ -2g(\widehat{h}_n, Z_i)+ \E_{D_n'} g(\widehat{h}_n, Z_i') \} \Big]  + 2\E_{D_n} \Big[\dfrac{1}{n}\sum_{i= 1}^n g(h_{\mathcal{H}_{\sigma, n}}, Z_i) \Big]
\\
 & \leq  \E_{D_n} \Big[\dfrac{1}{n}\sum_{i= 1}^n \{ -2g(\widehat{h}_n, Z_i)+ \E_{D_n'} g(\widehat{h}_n, Z_i') \} \Big] + 2 [R(h_{\mathcal{H}_{\sigma, n}}) - R(h^{*})]. 
\end{align}
%
%
%
The rest of the proof will consist of bounding $\E_{D_n} \Big[\dfrac{1}{n}\sum_{i= 1}^n \{ -2g(\widehat{h}_n, Z_i)+ \E_{D_n'} g(\widehat{h}_n, Z_i') \} \Big]$, $[R(h_{\mathcal{H}_{\sigma, n}}) - R(h^{*})]$ and then, derive a bound of the expected excess risk .

\medskip

\noindent
\textbf{Step 1 : Bounding the first term in the right-hand side of (\ref{Exp_excess_risk_v2})}.
%
%
 Let us denote for any $h \in \mathcal{H}_{\sigma, n}$
\[G(h, Z_i) =  \E_{D_n'} g(h, Z_i') -2g(h, Z_i).
\]
%
%
%
 Let $ \epsilon >0$. 
 Since $\|h\|_\infty \leq F_n < \infty$ for all $h \in \mathcal{H}_{\sigma, n}$, then, the $\epsilon$-covering number (see (\ref{epsi_covering_number}) in Section (\ref{asump})) of $\mathcal{H}_{\sigma, n}$ is finite. Set,
 \begin{equation}\label{proof_def_m_n}
 m := \mathcal{N}\Big( \mathcal{H}_{\sigma, n}, \epsilon  \Big).
\end{equation}    
From the Proposition 1 in \cite{ohn2019smooth}, we have,
\begin{equation}\label{proof_ing_covering_num}
\mathcal{N} \Big (\mathcal{H}_{\sigma, n}, \epsilon \Big)  \leq  \exp \Bigg (  2 L_n (S_n + 1) \log \left(\frac{1}{\epsilon} C_{\sigma} L_n (N_n + 1)(B_n \lor 
 1) \right) \Bigg),
\end{equation}
where $B_n \lor 1 = \max(B_n, 1)$.
The dependence of $m$ on $n$ is omitted to simplify notation.
Let $h_1,\cdots,h_m \in \mathcal{H}_{\sigma,n}$ such that,
\begin{equation}\label{proof_H_sigma_n_subset_ball}
\mathcal{H}_{\sigma,n} \subset \bigcup_{j=1}^m B(h_j, \epsilon).
\end{equation}
Recall that, $B(h_j, \epsilon)$ represents the ball of radius $\epsilon$, centered at $h_j$.  
As a consequence, there exists a random index $j^* \in \{1,\cdots, m\}$  such that, $ \|\widehat{h}_n - h_{j^*}\|_{\infty} \leq \epsilon$.
%
%
%
%
Recall that  $ g(h, Z_i) = \ell(h(X_i), Y_i) - \ell(h^{*}(X_i), Y_i)$.
Under the assumption (\textbf{A2}), we have for all $i=1,\cdots,n$, 
\begin{align*}
\nonumber |g(\widehat{h}_n), Z_i) - g(h_{j^*}, Z_i)| & = |\ell(\widehat{h}_n(X_i), Y_i) - \ell(h_{j^*}(X_i), Y_i)|  \leq \mathcal{K}_{\ell} |\widehat{h}_n(X_i) - h_{j^*}(X_i)| \leq \mathcal{K}_{\ell} \epsilon.
\end{align*}
Hence, 
\begin{align*}
\nonumber \dfrac{1}{n}\sum_{i=1}^n\E_{D_n}[g(\widehat{h}_n, Z_i)] & \leq  \dfrac{1}{n}\sum_{i=1}^n\E_{D_n}[g(h_{j^*}, Z_i)] + \dfrac{1}{n}\sum_{i=1}^n\E_{D_n}[g(\widehat{h}_n, Z_i) - g(h_{j^*}, Z_i)]
\\
\nonumber & \leq  \dfrac{1}{n}\sum_{i=1}^n\E_{D_n}[g(h_{j^*}, Z_i)]  +  \dfrac{1}{n}\sum_{i=1}^n\E_{D_n}[ |g(\widehat{h}_n, Z_i) - g(h_{j^*}, Z_i)|] 
\\
\nonumber & \leq  \dfrac{1}{n}\sum_{i=1}^n\E_{D_n}[g(h_{j^*}, Z_i)] + \mathcal{K}_{\ell} \epsilon.
\end{align*}
Thus, we have,
\begin{align*}
\nonumber |G(\widehat{h}_n, Z_i) - G(h_{j^*}, Z_i)| & = |\E_{D_n'} g(\widehat{h}_n, Z_i') -2g(\widehat{h}_n, Z_i) - \E_{D_n'} g(h_{j^*}, Z_i') + 2g(h_{j^*}, Z_i) |
\\
\nonumber & \leq  |\E_{D_n'} g(\widehat{h}_n, Z_i') - \E_{D_n'} g(h_{j^*}, Z_i') | + 2 |g(\widehat{h}_n, Z_i) - g(h_{j^*}, Z_i)|  \leq 3 \mathcal{K}_{\ell} \epsilon.
\end{align*}
Therefore,
\begin{equation}\label{euqua_bound_estimation_error}
 \E_{D_n} \Big[\dfrac{1}{n}\sum_{i= 1}^n \{ -2g(\widehat{h}_n, Z_i)+ \E_{D_n'} g(\widehat{h}_n, Z_i') \} \Big] = \E_{D_n}\big[\dfrac{1}{n}\sum_{i=1}^n G(\widehat{h}_n, Z_i) \big] \leq \E_{D_n}\big[\dfrac{1}{n}\sum_{i=1}^n G(h_{j^*}, Z_i) \big] + 3 \mathcal{K}_{\ell} \epsilon.   
\end{equation}

\medskip

For all $n \geq 1$, set:
\begin{equation}\label{proof_def_beta_n}
 \beta_n = \max\big(F_n, \big(n^{(\alpha)} \big)^{1/r} \big),
\end{equation}
where $r$ is given in the assumption \textbf{(A5)}.
%
%
 Let $T_{\beta_n}$ be a truncation at level $ \beta_n$, i.e., for any $ Y \in \R$,
\[
T_{\beta_n}Y =
\left\{
\begin{array}{ll}
 Y    &  \text{if} ~ |Y| \leq \beta_n\\
  \beta_n \cdot \text{sign}(Y)   &  \text{otherwise}.
\end{array}
\right.
\]
Let us define the function $h_{\beta_n}^{*} : \mathcal{X} \rightarrow \R$ by 
\begin{equation}\label{def_h_beta_n}
h_{\beta_n}^{*}(x) = \underset{h(x): \|h\|_{\infty}\leq \beta_n}{\argmin} \E[\ell(h(X), T_{\beta_n}Y)|X=x],
\end{equation}
for each $x \in \mathcal{X}$. 
Let us recall that $\|h^{*}\|_{\infty} \leq \mathcal{K} \leq F_n \leq \beta_n$ and 
\[h^{*}(x) = \underset{h(x): \|h\|_{\infty} \leq \beta_n}{\argmin} \E[\ell(h(X), Y)|X=x].
\]
Thus, for any $h$ satisfying $\|h\|_{\infty} \leq \beta_n$, we have from the definition above,
\begin{equation}\label{proof_ing_ell_T_beta}
\E[\ell(h_{\beta_n}^{*}(X_i), T_{\beta_n}Y_i)] \leq \E[\ell(h(X_i), T_{\beta_n}Y_i)] ~ \text{and} ~  \E[\ell(h^{*}(X_i), Y_i)] \leq E[\ell(h(X_i), Y_i)]. 
\end{equation}
For all $ h \in \mathcal{H}_{\sigma, n}$, set 

\begin{equation}\label{def_g_beta}
 g_{\beta_n}(h(X_i), Z_i) = \ell(h(X_i), T_{\beta_n}Y_i) - \ell(h_{\beta_n}^{*}(X_i), T_{\beta_n}Y_i).
\end{equation}
We get,
\begin{align}
\nonumber \E[g(h, Z_i)] & = \E\Big[g_{\beta_n}(h, Z_i) - g_{\beta_n}(h, Z_i) + \ell(h(X_i), Y_i) - \ell(h^{*}(X_i), Y_i) \\
\nonumber & \hspace{3cm} + \ell(h_{\beta_n}^{*}(X_i), T_{\beta_n}Y_i) - \ell(h_{\beta_n}^{*}(X_i), T_{\beta_n}Y_i) \Big]
\\
\nonumber & \leq \E\Big[g_{\beta_n}(h, Z_i) - \ell(h(X_i), T_{\beta_n}Y_i)  + \ell(h_{\beta_n}^{*}(X_i), T_{\beta_n}Y_i) + \ell(h(X_i), Y_i) - \ell(h^{*}(X_i), Y_i) \\
\nonumber & \hspace{3cm} + 
\ell(h^{*}(X_i), T_{\beta_n}Y_i) - \ell(h^{*}(X_i), T_{\beta_n}Y_i)\Big]
\\
\nonumber & \leq \E[g_{\beta_n}(h, Z_i)] + \E[\ell(h(X_i), Y_i) - \ell(h(X_i), T_{\beta_n}Y_i) + \ell(h^{*}(X_i), T_{\beta_n}Y_i) - \ell(h^{*}(X_i), Y_i)\\
\nonumber & \hspace{1cm}+ \ell(h_{\beta_n}^{*}(X_i), T_{\beta_n}Y_i) - \ell(h^{*}(X_i), T_{\beta_n}Y_i)]
\\
\label{proof_g_beta_h_ell_h} & \leq  \E[g_{\beta_n}(h, Z_i)] + \E[\ell(h(X_i), Y_i) - \ell(h(X_i), T_{\beta_n}Y_i) + \ell(h^{*}(X_i), T_{\beta_n}Y_i) - \ell(h^{*}(X_i), Y_i)]
\\
\label{proof_g_beta_h_2K_ell} & \leq \E[g_{\beta_n}(h, Z_i)] + 2\mathcal{K}_{\ell}\E[|T_{\beta_n}Y_i - Y_i|],
\end{align}
where the inequality in (\ref{proof_g_beta_h_ell_h}) holds from (\ref{proof_ing_ell_T_beta}).
Moreover, we have
\begin{align}\label{proof_T_beta_Yi}
\nonumber |T_{\beta_n}Y_i - Y_i| & = |\beta_n \cdot \text{sign}(Y_i)\ind_{\{|Y_i| \ge \beta_n\}} + Y_i\ind_{\{|Y_i| \leq \beta_n\}} - Y_i|
 \leq |Y_i|\ind_{\{|Y_i| \ge \beta_n\}}.
\end{align}
Recall the assumption \textbf{(A5)} that $Y_i$ has a finite $r$-moment, for some $r >1$. We have,
\begin{equation}\label{proof_Y_i_abs_beta_n}
  \E[|Y_i|\ind_{\{|Y_i| \ge \beta_n\}}]  \leq \E[|Y_i||Y_i|^{r - 1}]/\beta_n^{r - 1}]  \leq \E[|Y_i|^r]/\beta_n^{r - 1}. 
\end{equation}
Therefore, from (\ref{proof_g_beta_h_2K_ell}) and (\ref{proof_Y_i_abs_beta_n}), we get
\begin{equation}\label{proof_g_h_ing_g_beta}
\E[g(h, Z_i)] \leq \E[g_{\beta_n}(h, Z_i)] + 2 \mathcal{K}_{\ell}\E[|Y_i|^r]/\beta_n^{r - 1}.
\end{equation}
Using a similar argument, we have
\begin{align}\label{proof_g_beta_ing_g_h}
\nonumber  \E[g_{\beta_n}(h, Z_i)] & = \E[g(h, Z_i) - g(h, Z_i) + g_{\beta_n}(h, Z_i)]
\\
\nonumber & =  \E[g(h, Z_i)] + \E[ \ell(h^{*}(X_i), Y_i) - \ell(h(X_i), Y_i) + \ell(h(X_i), T_{\beta_n}Y_i) - \ell(h_{\beta_n}^{*}(X_i), T_{\beta_n}Y_i) \\
\nonumber & \hspace{1cm} + \ell(h_{\beta_n}^{*}(X_i), Y_i) - \ell(h_{\beta_n}^{*}(X_i),Y_i)] 
\\
\nonumber & \leq  \E[g(h, Z_i)] + \E[ \ell(h^{*}(X_i), Y_i) -  \ell(h_{\beta_n}^{*}(X_i),Y_i) + \ell(h(X_i), T_{\beta_n}Y_i)  - \ell(h(X_i), Y_i)   \\
\nonumber & \hspace{3cm} + \ell(h_{\beta_n}^{*}(X_i), Y_i) -\ell(h_{\beta_n}^{*}(X_i), T_{\beta_n}Y_i)]  
\\
\nonumber & \leq  \E[g(h, Z_i)] + \E[\ell(h(X_i), T_{\beta_n}Y_i)  - \ell(h(X_i), Y_i) + \ell(h_{\beta_n}^{*}(X_i), Y_i) -\ell(h_{\beta_n}^{*}(X_i), T_{\beta_n}Y_i)] 
\\
\nonumber & \leq \E[g(h, Z_i)] + 2\mathcal{K}_{\ell}\E[| T_{\beta_n}Y_i -Y_i|]
\\
& \leq  \E[g(h, Z_i)] + 2\mathcal{K}_{\ell} \E[|Y_i|^r]/\beta_n^{r - 1}.
\end{align}
Note that, $g(h, Z_i')$ and $g_{\beta_n}(h, Z_i')$ also satisfy the above inequalities.
Let $ h \in \mathcal{H}_{\sigma, n}$. 
Set 
\begin{equation}\label{def_G_beta}
 G_{\beta_n}(h, Z_i) := \E_{D_n'}[g_{\beta_n}(h, Z_i')] - 2g_{\beta_n}(h, Z_i). 
\end{equation}
We have
\begin{align*}
\nonumber |G(h, Z_i) - G_{\beta_n}(h, Z_i)| & = | \E_{D_n'} g(h, Z_i') -2g(h, Z_i) -\E_{D_n'}[g_{\beta_n}(h, Z_i)] + 2g_{\beta_n}(h, Z_i)|
\\
& \leq | \E_{D_n'} g(h, Z_i') - \E_{D_n'}[g_{\beta_n}(h, Z_i')] | + 2|g(h, Z_i) - g_{\beta_n}(h, Z_i)|.
\end{align*}
Thus, in addition to (\ref{proof_g_h_ing_g_beta}) and (\ref{proof_g_beta_ing_g_h}), It comes that
\begin{equation}\label{Exp_bound_Gbeta}
\E_{D_n} \Big[ \dfrac{1}{n} \sum_{i=1}^n G(h, Z_i) \Big] \leq \E_{D_n}\Big[ \dfrac{1}{n} \sum_{i = 1}^n G_{\beta_n}(h, Z_i) \Big] + 6\mathcal{K}_{\ell} \E[|Y_i|^r]/ \beta_n^{r-1}. 
\end{equation}
Therefore, in addition to (\ref{euqua_bound_estimation_error}), we get,
\begin{align}\label{proof_exp_2_g_ing} 
 \nonumber \E_{D_n} \Big[\dfrac{1}{n}\sum_{i= 1}^n \{ -2g(\widehat{h}_n, Z_i)+ \E_{D_n'} g(\widehat{h}_n, Z_i') \} \Big] &  \leq \E_{D_n}\big[\dfrac{1}{n}\sum_{i=1}^n G(h_{j^*}, Z_i) \big]  + 3 \mathcal{K}_{\ell} \epsilon \\
 & \leq \E_{D_n}\Big[ \dfrac{1}{n} \sum_{i = 1}^n G_{\beta_n}(h_{j^*}, Z_i) \Big] +  6\mathcal{K}_{\ell} \E[|Y_i|^r]/ \beta_n^{r-1} + 3\mathcal{K}_{\ell}\epsilon.     
\end{align} 
Let us focus on an upper bound of $\E_{D_n}\Big[\dfrac{1}{n} \sum_{i = 1}^n G_{\beta_n}(h_{j^*}, Z_i) \Big]$.
%
%
%
Let $ h \in \mathcal{H}_{\sigma, n}$. Since $F_n \leq \beta_n$, we have,
\begin{align}\label{proof_g_beta_K_ell_beta}
  |g_{\beta_n}(h, Z_i)| & = |\ell(h(X_i), T_{\beta_n} Y_i) - \ell(h_{\beta_n}^{*}(X_i), T_{\beta_n} Y_i)| \leq \mathcal{K}_{\ell} | h(X_i) - h_{\beta_n}^{*}(X_i)| \leq \mathcal{K}_{\ell}(F_n +\beta_n) \leq 2\mathcal{K}_{\ell}\beta_n.
\end{align}
 According to (\ref{proof_g_beta_K_ell_beta}), we have,
\begin{align}\label{equa_bound_of_the_variance}
  \var(g_{\beta_n}(h, Z_i))  \leq \E[g_{\beta_n}(h, Z_i)^2]  \leq \E[|g_{\beta_n}(h, Z_i)||g_{\beta_n}(h, Z_i)|]
  \leq  2\mathcal{K}_{\ell}\beta_n \E[g_{\beta_n}(h, Z_i)]. 
\end{align}
Let $ \varepsilon > 0$ and $n^{(\alpha)} = \lfloor n \lceil \{8n/c\}^{1/(\gamma + 1)}\rceil^{-1} \rfloor \ge 2$ for some $ \overline{\alpha} > 0, \gamma > 0$ given in the assumption \textbf{(A4)}. According to (\ref{proof_g_beta_K_ell_beta}), we have, $|\E_{D_n'}[g_{\beta_n}(h_{j^*}, Z_i')] - g_{\beta_n}(h_{j^*}, Z_i)| \leq 4\mathcal{K}_{\ell} \beta_n$. By using (\ref{equa_bound_of_the_variance}) and the Bernstein type inequality for dependent data (see (\cite{zou2009generalization}, \cite{modha1996minimum})), we have for all $j=1,\cdots,m$ (where $m$ is defined in (\ref{proof_def_m_n})),
\begin{align}\label{term_bound_excess_risk}
\nonumber P \Bigg\{\dfrac{1}{n} \sum_{i=1}^n G_{\beta_n}(h_{j}, Z_i) > \varepsilon \Bigg\}  & = P \Bigg\{\dfrac{1}{n} \sum_{i=1}^n\E_{D_n'}[g_{\beta_n}(h_{j}, Z_i')] - \dfrac{2}{n} \sum_{i = 1}^n g_{\beta_n}(h_{j}, Z_i) > \varepsilon \Bigg\}
\\
\nonumber & =  P \Big\{\E_{D_n'}[g_{\beta_n}(h_{j}, Z_1')] - \dfrac{2}{n} \sum_{i = 1}^n g_{\beta_n}(h_{j}, Z_i) > \varepsilon \Big\}.
\\
\nonumber & =  P \Big\{2\E_{D_n'}[g_{\beta_n}(h_{j}, Z_1')] - \dfrac{2}{n} \sum_{i = 1}^n g_{\beta_n}(h_{j^*}, Z_i) > \varepsilon + \E_{D_n'}[g_{\beta_n}(h_{j}, Z_1')] \Big\}
\\
\nonumber & =  P \Big\{\E_{D_n'}[g_{\beta_n}(h_{j}, Z_1')] - \dfrac{1}{n} \sum_{i = 1}^n g_{\beta_n}(h_{j}, Z_i) > \dfrac{\varepsilon}{2} + \dfrac{1}{2} \E_{D_n'}[g_{\beta_n}(h_{j^*}, Z_1')]\Big\}
\\
\nonumber & \leq  P \Bigg\{\E_{D_n'}[g_{\beta_n}(h_{j}, Z_1')] - \dfrac{1}{n} \sum_{i = 1}^n g_{\beta_n}(h_{j}, Z_i) > \dfrac{\varepsilon}{2} + \dfrac{\E_{D_n'}[|g_{\beta_n}(h_{j}, Z_1')|^2]}{4 \mathcal{K}_{\ell}\beta_n} \Bigg\}
\\
\nonumber & \leq  P \Bigg\{\dfrac{1}{n} \sum_{i = 1}^n \Big(\E[g_{\beta_n}(h_{j}, Z_i)] - g_{\beta_n}(h_{j}, Z_i)\Big) > \dfrac{\varepsilon}{2} + \dfrac{\E[|g_{\beta_n}(h_{j}, Z_1)|^2]}{4 \mathcal{K}_{\ell}\beta_n} \Bigg\}
\\
\nonumber & \leq (1 + 4e^{-2}\overline{\alpha})\exp\Bigg(-\dfrac{\Big(\dfrac{\varepsilon}{2} + \dfrac{\E[|g_{\beta_n}(h_{j}, Z_1)|^2]}{ 4\mathcal{K}_{\ell}\beta_n} \Big)^2 n^{(\alpha)}}{2 \Big(\E[|g_{\beta_n}(h_{j}, Z_1)|^2] +  4\Big(\dfrac{\varepsilon}{2} + \dfrac{\E[|g_{\beta_n}(h_{j}, Z_1)|^2]}{4\mathcal{K}_{\ell}\beta_n} \Big)\mathcal{K}_{\ell} \beta_n/3 \Big)} \Bigg)
\\
\nonumber & \leq (1 + 4e^{-2}\overline{\alpha})\exp\Bigg(-\dfrac{\Big(\dfrac{\varepsilon}{2} + \dfrac{\E[|g_{\beta_n}(h_{j}, Z_1)|^2]}{4 \mathcal{K}_{\ell}\beta_n} \Big)^2 n^{(\alpha)}}{2 \Big(\Big(\dfrac{\varepsilon}{2} + \dfrac{\E[|g_{\beta_n}(h_{j}, Z_1)|^2]}{4\mathcal{K}_{\ell}\beta_n} \Big) 4\mathcal{K}_{\ell} \beta_n +  4\Big(\dfrac{\varepsilon}{2} + \dfrac{\E[|g_{\beta_n}(h_{j}, Z_1)|^2]}{4 \mathcal{K}_{\ell}\beta_n} \Big)\mathcal{K}_{\ell} \beta_n/3 \Big)} \Bigg)
\\
\nonumber & \leq (1 + 4e^{-2}\overline{\alpha})\exp\Bigg(-3\dfrac{\Big(\dfrac{\varepsilon}{2} + \dfrac{\E[|g_{\beta_n}(h_{j}, Z_1)|^2]}{4\mathcal{K}_{\ell}\beta_n} \Big)^2 n^{(\alpha)}}{32\Big(\dfrac{\varepsilon}{2} + \dfrac{\E[|g_{\beta_n}(h_{j}, Z_1)|^2]}{ 4\mathcal{K}_{\ell}\beta_n} \Big)\mathcal{K}_{\ell} \beta_n} \Bigg)
\\
\nonumber & \leq (1 + 4e^{-2}\overline{\alpha})\exp\Bigg(-3\dfrac{\Big(\dfrac{\varepsilon}{2} + \dfrac{\E[|g_{\beta_n}(h_{j}, Z_1)|^2]}{4\mathcal{K}_{\ell}\beta_n} \Big) n^{(\alpha)}}{32\mathcal{K}_{\ell} \beta_n} \Bigg)
\\
%
& \leq (1 + 4e^{-2}\overline{\alpha})\exp\Bigg(-\dfrac{ 3\varepsilon   }{64\mathcal{K}_{\ell} \beta_n}  n^{(\alpha)}\Bigg) .
\end{align}
Set, $\epsilon = \frac{1}{n^{(\alpha)}}$, see (\ref{proof_def_m_n}). That is $m = \mathcal{N}\Big( \mathcal{H}_{\sigma, n},  1/ n^{(\alpha)} \Big)$. In addition to (\ref{proof_ing_covering_num}) (see the proof of Theorem \ref{thm1}) we have,
\begin{align}
\nonumber P \Big\{\dfrac{1}{n} \sum_{i = 1}^n G_{\beta_n}(h_{j^*}, Z_i) > \varepsilon \Big\}  & \leq   P \Bigg\{ \bigcup_{j=1}^m \Big( \dfrac{1}{n} \sum_{i = 1}^n G_{\beta_n}(h_{j}, Z_i) > \varepsilon \Big) \Bigg\} \leq \sum_{j=1}^m  P \Big\{\dfrac{1}{n} \sum_{i = 1}^n G_{\beta_n}(h_{j}, Z_i) > \varepsilon \Big\}   \\
& \leq \mathcal{N}\Big( \mathcal{H}_{\sigma, n}, \frac{1}{n^{(\alpha)}}\Big)\cdot (1 + 4e^{-2}\overline{\alpha})\exp\Bigg(-\dfrac{ 3\varepsilon   }{64\mathcal{K}_{\ell} \beta_n}  n^{(\alpha)} \Bigg)
\\
\nonumber & \leq (1 + 4e^{-2}\overline{\alpha})\exp \Bigg(2L_n(S_n + 1)\log\Big(n^{(\alpha)}C_{\sigma}L_n(N_n + 1)(B_n \lor 1)\Big) -\dfrac{ 3\varepsilon   }{64\mathcal{K}_{\ell} \beta_n}  n^{(\alpha)} \Bigg). 
\end{align}
For $\alpha_n > 0$, we have
\begin{align}
\nonumber  \E\Big[\dfrac{1}{n}\sum_{i = 1}^nG_{\beta_n}(h_{j^*}, Z_i) \Big] & \leq \alpha_n + \displaystyle\int_{\alpha_n}^{\infty} P\Big\{\dfrac{1}{n}\sum_{i = 1}^nG_{\beta_n}(h_{j^*}, Z_i) > \varepsilon\Big\} d\varepsilon  
\\
\nonumber & \leq \alpha_n + (1 + 4e^{-2}\overline{\alpha}) \exp \Big(2L_n(S_n + 1)\log(n^{(\alpha)} C_{\sigma}L_n(N_n + 1)(B_n \lor 1)) \Big)  \times \displaystyle\int_{\alpha_n}^{\infty} \exp\Big(-\dfrac{3\varepsilon  n^{(\alpha)}}{64\mathcal{K}_{\ell} \beta_n}\Big) d\varepsilon.    
\end{align}
One can easily see that,
\begin{equation}
 \displaystyle\int_{\alpha_n}^{\infty} \exp\Big(-\dfrac{3\varepsilon n^{(\alpha)}}{64\mathcal{K}_{\ell}\beta_n}\Big) d\varepsilon = \frac{64\mathcal{K}_{\ell} \beta_n}{3n^{(\alpha)}}\exp\Big(-\dfrac{3\alpha_n n^{(\alpha)}}{64\mathcal{K}_{\ell}\beta_n}\Big)
\end{equation}
Hence we have,
\begin{align}\label{equa1_bound_G_beta_n}
\nonumber & \E\Big[\dfrac{1}{n}\sum_{i = 1}^nG_{\beta_n}(h_{j^*}, Z_i) \Big]  \leq \alpha_n + \displaystyle\int_{\alpha_n}^{\infty} P\Big\{\dfrac{1}{n}\sum_{i = 1}^nG_{\beta_n}(h_{j^*}, Z_i) > \varepsilon\Big\} d\varepsilon  
\\
\nonumber & \leq \alpha_n +
\frac{64\mathcal{K}_{\ell} \beta_n (1 + 4e^{-2}\overline{\alpha})}{3n^{(\alpha)}}\exp\Big(2L_n(S_n + 1)\log(n^{(\alpha)} C_{\sigma}L_n(N_n + 1)(B_n \lor 1))  \Big) \exp\Big(-\dfrac{3\alpha_n n^{(\alpha)}}{64\mathcal{K}_{\ell}\beta_n}\Big)
\\
 & \leq  \alpha_n + \frac{64\mathcal{K}_{\ell} \beta_n (1 + 4e^{-2}\overline{\alpha})}{3n^{(\alpha)}}\exp\Big(2L_n(S_n + 1)\log(n^{(\alpha)}C_{\sigma}L_n(N_n + 1)(B_n \lor 1)) - \dfrac{3\alpha_n n^{(\alpha)}}{64\mathcal{K}_{\ell}\beta_n} \Big).
\end{align}
Set $ \alpha_n \coloneqq \dfrac{ \big(\log n^{(\alpha)}  \big)^{\nu}}{ \big( n^{(\alpha)} \big)^{\frac{s}{s + d}(1-\frac{1}{r}) } }$, for some $\nu >3$. According to (\ref{equa1_bound_G_beta_n}) and since $\beta_n \geq \big(n^{(\alpha)} \big)^{1/r}$ (see (\ref{proof_def_beta_n})), we have 
\begin{align}\label{proof_E_G_beta_log_nu_n}
\nonumber &\E\Big[ \dfrac{1}{n}\sum_{i = 1}^nG_{\beta_n}(h_{j^*}, Z_i) \Big] \\
\nonumber & \leq \dfrac{ \big(\log n^{(\alpha)}  \big)^{\nu}}{ \big( n^{(\alpha)} \big)^{\frac{s}{s + d}(1-\frac{1}{r}) } }  + \frac{64\mathcal{K}_{\ell} \beta_n (1 + 4e^{-2}\overline{\alpha})}{3n^{(\alpha)}}\exp\Big(2L_n(S_n + 1)\log(n^{(\alpha)}C_{\sigma}L_n(N_n + 1)(B_n \lor 1))  - \dfrac{3  \big(n^{(\alpha)}\big)^{d/(s+d)} \big(\log n^{(\alpha)}  \big)^{\nu}  }{64\mathcal{K}_{\ell}\beta_n} \Big) \\
& \leq \dfrac{ \big(\log n^{(\alpha)}  \big)^{\nu}}{ \big( n^{(\alpha)} \big)^{\frac{s}{s + d}(1-\frac{1}{r}) } }  + \frac{64\mathcal{K}_{\ell}  (1 + 4e^{-2}\overline{\alpha})}{3 \big(n^{(\alpha)} \big)^{(1-1/r)} }\exp\Big(2L_n(S_n + 1)\log(n^{(\alpha)}C_{\sigma}L_n(N_n + 1)(B_n \lor 1))  - \dfrac{3  \big(n^{(\alpha)}\big)^{ (1-\frac{1}{r}) \frac{d}{s+d} } \big(\log n^{(\alpha)}  \big)^{\nu}  }{64\mathcal{K}_{\ell} } \Big).
\end{align}
Recall the assumptions:
$
L_n = \big(1 - \frac{1}{r} \big) \dfrac{s L_0}{s + d} \log\big(n^{(\alpha)} \big), N_n = N_0 \big(n^{(\alpha)} \big)^{\big(1 - \frac{1}{r} \big) \frac{d}{s + d}}, S_n = \big(1 - \frac{1}{r} \big) \frac{s S_0}{s + d} (n^{(\alpha)})^{\big(1 - \frac{1}{r} \big) \frac{d}{s + d}}\log\big(n^{(\alpha)} \big), 
$
$ B_n = B_0 \big(n^{(\alpha)} \big)^{ \big(1 - \frac{1}{r} \big) \frac{4s(d/s + 1)}{s + d}}$,
with $L_0, N_0, B_0, S_0 >0$. 
Hence,
\begin{align*}
\nonumber &\E\Big[\dfrac{1}{n}\sum_{i = 1}^nG_{\beta_n}(h_{j^*}, Z_i) \Big] \\
\nonumber & \leq \dfrac{ \big(\log n^{(\alpha)}  \big)^{\nu}}{ \big( n^{(\alpha)} \big)^{\frac{s}{s + d}(1-\frac{1}{r}) } }   + \frac{64\mathcal{K}_{\ell}  (1 + 4e^{-2}\overline{\alpha})}{3 \big(n^{(\alpha)} \big)^{(1-1/r)} }\exp\Bigg[2\big(1 - \frac{1}{r} \big) \dfrac{s L_0}{s + d} \log\big(n^{(\alpha)} \big) \Big(\big(1 - \frac{1}{r} \big) \frac{s S_0}{s + d} (n^{(\alpha)})^{\big(1 - \frac{1}{r} \big) \frac{d}{s + d}}\log\big(n^{(\alpha)} \big) + 1 \Big) \\ 
\nonumber & \hspace{3.5cm} \times \log\Bigg(n^{(\alpha)}C_{\sigma} \big(1 - \frac{1}{r} \big) \dfrac{s L_0}{s + d} \log\big(n^{(\alpha)} \big) \big(N_0 \big(n^{(\alpha)} \big)^{\big(1 - \frac{1}{r} \big) \frac{d}{s + d}} + 1 \big) \big(B_0 \big(n^{(\alpha)} \big)^{ \big(1 - \frac{1}{r} \big) \frac{4s(d/s + 1)}{s + d}} \lor 1 \big) \Bigg) \\
\nonumber &  \hspace{12.8cm} - \dfrac{3  \big(n^{(\alpha)}\big)^{ (1-\frac{1}{r}) \frac{d}{s+d} } \big(\log n^{(\alpha)}  \big)^{\nu}  }{64\mathcal{K}_{\ell} } \Bigg] 
\end{align*}
\begin{align}\label{equa2_bound_G_beta_n}
\nonumber &  \leq \dfrac{ \big(\log n^{(\alpha)}  \big)^{\nu}}{ \big( n^{(\alpha)} \big)^{\frac{s}{s + d}(1-\frac{1}{r}) } }   + \frac{64\mathcal{K}_{\ell}  (1 + 4e^{-2}\overline{\alpha})}{3 \big(n^{(\alpha)} \big)^{(1-1/r)} }\exp\Bigg[\dfrac{3  \big(n^{(\alpha)}\big)^{ (1-\frac{1}{r}) \frac{d}{s+d} } \big(\log n^{(\alpha)}  \big)^{\nu}  }{64\mathcal{K}_{\ell} } \\
\nonumber & \hspace{8cm} \times  \dfrac{128\mathcal{K}_{\ell} \big(1 - \frac{1}{r} \big) \dfrac{s L_0}{s + d} \log\big(n^{(\alpha)} \big) \Big(\big(1 - \frac{1}{r} \big) \frac{s S_0}{s + d}\log\big(n^{(\alpha)} \big) + 1 \Big)}{3 \big(\log n^{(\alpha)}  \big)^{\nu}} \\
& \hspace{2cm}   \times \Bigg( \log\Big(n^{(\alpha)}C_{\sigma} \big(1 - \frac{1}{r} \big) \dfrac{s L_0}{s + d} \log\big(n^{(\alpha)} \big) \big(N_0 \big(n^{(\alpha)} \big)^{\big(1 - \frac{1}{r} \big) \frac{d}{s + d}} + 1 \big) \big(B_0 \big(n^{(\alpha)} \big)^{ \big(1 - \frac{1}{r} \big) \frac{4s(d/s + 1)}{s + d}} \lor 1 \big) \Big)
- 1 \Bigg) \Bigg].
\end{align}

\medskip


\medskip

\noindent
%
%
From (\ref{equa2_bound_G_beta_n}), we have
\begin{align*}
& \dfrac{128\mathcal{K}_{\ell} \big(1 - \frac{1}{r} \big) \dfrac{s L_0}{s + d} \log\big(n^{(\alpha)} \big) \Big(\big(1 - \frac{1}{r} \big) \frac{s S_0}{s + d}\log\big(n^{(\alpha)} \big) + 1 \Big)}{3 \big(\log n^{(\alpha)}  \big)^{\nu}} \\
 & \times \log\Big(n^{(\alpha)}C_{\sigma} \big(1 - \frac{1}{r} \big) \dfrac{s L_0}{s + d} \log\big(n^{(\alpha)} \big) \big(N_0 \big(n^{(\alpha)} \big)^{\big(1 - \frac{1}{r} \big) \frac{d}{s + d}} + 1 \big) \big(B_0 \big(n^{(\alpha)} \big)^{ \big(1 - \frac{1}{r} \big) \frac{4s(d/s + 1)}{s + d}} \lor 1 \big) \Big) \limiten 0.
\end{align*}
Hence, there exists $n_0 = n_0(\mathcal{K}_{\ell}, L_0, N_0, B_0, S_0, C_\sigma, s, d, r, c, \gamma)$ such that, for any $n \ge n_0$, we have
\begin{align}\label{equa_limite}
\nonumber & \dfrac{128\mathcal{K}_{\ell} \big(1 - \frac{1}{r} \big) \dfrac{s L_0}{s + d} \log\big(n^{(\alpha)} \big) \Big(\big(1 - \frac{1}{r} \big) \frac{s S_0}{s + d}\log\big(n^{(\alpha)} \big) + 1 \Big)}{3 \big(\log n^{(\alpha)}  \big)^{\nu}} \\
 & \times \log\Big(n^{(\alpha)}C_{\sigma} \big(1 - \frac{1}{r} \big) \dfrac{s L_0}{s + d} \log\big(n^{(\alpha)} \big) \big(N_0 \big(n^{(\alpha)} \big)^{\big(1 - \frac{1}{r} \big) \frac{d}{s + d}} + 1 \big) \big(B_0 \big(n^{(\alpha)} \big)^{ \big(1 - \frac{1}{r} \big) \frac{4s(d/s + 1)}{s + d}} \lor 1 \big) \Big) < \frac{1}{2}.
\end{align}
According to (\ref{equa_limite}), we have
\begin{align*}
\E\Big[\dfrac{1}{n}\sum_{i = 1}^nG_{\beta_n}(h_{j^*}, Z_i) \Big] & \leq \dfrac{ \big(\log n^{(\alpha)}  \big)^{\nu}}{ \big( n^{(\alpha)} \big)^{\frac{s}{s + d}(1-\frac{1}{r}) } }   + \frac{64\mathcal{K}_{\ell}  (1 + 4e^{-2}\overline{\alpha})}{3 \big(n^{(\alpha)} \big)^{(1-1/r)} } \exp\Big(- \dfrac{3  \big(n^{(\alpha)}\big)^{ (1-\frac{1}{r}) \frac{d}{s+d} } \big(\log n^{(\alpha)}  \big)^{\nu}  }{128\mathcal{K}_{\ell} } \Big) \\
&\leq \dfrac{ \big(\log n^{(\alpha)}  \big)^{\nu}}{ \big( n^{(\alpha)} \big)^{\frac{s}{s + d}(1-\frac{1}{r}) } }   + \frac{64\mathcal{K}_{\ell}  (1 + 4e^{-2}\overline{\alpha})}{3 \big(n^{(\alpha)} \big)^{(1-1/r)} }.
\end{align*}
%
%
So, (\ref{euqua_bound_estimation_error}) and (\ref{proof_exp_2_g_ing}) give,
\begin{align*}\label{equa_bound_stochastic_error}
\E_{D_n}  \Big[\dfrac{1}{n}\sum_{i= 1}^n \{ -2g(\widehat{h}_n, Z_i)+ \E_{D_n'} g(\widehat{h}_n, Z_i') \} \Big] & \leq  \dfrac{ \big(\log n^{(\alpha)}  \big)^{\nu}}{ \big( n^{(\alpha)} \big)^{\frac{s}{s + d}(1-\frac{1}{r}) } }   + \frac{64\mathcal{K}_{\ell}  (1 + 4e^{-2}\overline{\alpha})}{3 \big(n^{(\alpha)} \big)^{(1 - \frac{1}{r})} } + 6\mathcal{K}_{\ell} \E[|Y_i|^r]/ \beta_n^{r-1} + 3\mathcal{K}_{\ell} \epsilon \\
 & \leq \dfrac{ \big(\log n^{(\alpha)}  \big)^{\nu}}{ \big( n^{(\alpha)} \big)^{\frac{s}{s + d}(1-\frac{1}{r}) } }   + \frac{64\mathcal{K}_{\ell}  (1 + 4e^{-2}\overline{\alpha})}{3 \big(n^{(\alpha)} \big)^{(1 - \frac{1}{r})} }  +  \dfrac{6\mathcal{K}_{\ell} \E[|Y_i|^r]}{ (n^{(\alpha)})^{(1 - \frac{1}{r})}} + 3\mathcal{K}_{\ell} \epsilon.    
\end{align*} 
Recall that from assumption \textbf{(A4)}, $\E[|Y_0|^r] \leq M$ for some $M >0, r >1$.
Since we have set $\epsilon = \dfrac{1}{n^{(\alpha)}}$, it holds that,
\begin{equation} \label{equa_bound_stochastic_error_v2}
 \E_{D_n}  \Big[\dfrac{1}{n}\sum_{i= 1}^n \{ -2g(\widehat{h}_n, Z_i)+ \E_{D_n'} g(\widehat{h}_n, Z_i') \} \Big]  \leq \dfrac{ \big(\log n^{(\alpha)}  \big)^{\nu}}{ \big( n^{(\alpha)} \big)^{\frac{s}{s + d}(1-\frac{1}{r}) } }   + \frac{64\mathcal{K}_{\ell}  (1 + 4e^{-2}\overline{\alpha})}{3 \big(n^{(\alpha)} \big)^{(1 - \frac{1}{r})} }  +  \dfrac{6\mathcal{K}_{\ell} M}{ (n^{(\alpha)})^{(1 - \frac{1}{r})}} + \frac{3\mathcal{K}_{\ell}}{n^{(\alpha)}}.    
\end{equation} 

\medskip

\noindent
\textbf{Step 2 : Bounding the second term in the right-hand side of (\ref{Exp_excess_risk_v2})}. 
%

\medskip

\noindent
Recall that we have set $\mathcal{H}_{\sigma, n} \coloneqq \mathcal{H}_{\sigma}(L_n, N_n, B_n, F_n, S_n)$ ( see (\ref{proof_def_H_sigma_n})).

\medskip

\noindent
Under the Lipschitz property on $\ell$, for all $ h \in \mathcal{H}_{\sigma, n}$, the following inequality holds,
\begin{equation}\label{equa_excess_risk1}
R(h) - R(h^{*}) = \E_{Z_0} [\ell(h(X_0), Y_0)] - \E_{Z_0} [\ell(h^{*}(X_0), Y_0)] \leq \mathcal{K}_{\ell} \E_{X_0} [ |h(X_0) - h^{*} (X_0) |].
\end{equation}
According to (\ref{equa_excess_risk1}), we have,
\begin{align}\label{equa_approxi_error1}
\nonumber R(h_{\mathcal{H}_{\sigma, n}}) - R(h^{*}) & = \underset{h\in \mathcal{H}_{\sigma, n}}{\inf} R(h) - R(h^{*}) = \underset{h\in \mathcal{H}_{\sigma, n}}{\inf} \Big(R(h) - R(h^{*}) \Big)
\\
& \leq \mathcal{K}_{\ell} \underset{h\in \mathcal{H}_{\sigma, n}}{\inf}\E_{X_0} [ |h(X_0) - h^{*} (X_0) |].
\end{align}
Recall the assumption $h^{*} \in C^{s, \mathcal{K}}(\mathcal{X})$ for some $s, \mathcal{K} > 0$. 
Set $  \epsilon_n = \dfrac{ 1}{ \big( n^{(\alpha)} \big)^{\frac{s}{s + d}(1-\frac{1}{r}) } } $.
From \cite{kengne2023excess}, there exists a positive constants $L_0, N_0, B_0, S_0 >0$ such that with 
$L_n = \big(1 - \frac{1}{r} \big) \dfrac{s L_0}{s + d} \log\big(n^{(\alpha)} \big), N_n = N_0 \big(n^{(\alpha)} \big)^{\big(1 - \frac{1}{r} \big) \frac{d}{s + d}}, S_n = \big(1 - \frac{1}{r} \big) \frac{s S_0}{s + d} (n^{(\alpha)})^{\big(1 - \frac{1}{r} \big) \frac{d}{s + d}}\log\big(n^{(\alpha)} \big), 
$
$ B_n = B_0 \big(n^{(\alpha)} \big)^{ \big(1 - \frac{1}{r} \big) \frac{4s(d/s + 1)}{s + d}}$, there is a neural network $h_{n} \in \mathcal{H}_{\sigma, n}$ satisfying,
\begin{equation}\label{cond_excess_risk_bound}
\| h_{n} - h^{*} \|_{\infty, \mathcal{X}} < \dfrac{ 1}{ \big( n^{(\alpha)} \big)^{\frac{s}{s + d}(1-\frac{1}{r}) } }.
\end{equation}
According to (\ref{equa_approxi_error1}) and in addition to (\ref{cond_excess_risk_bound}), we have,
\begin{equation}\label{approxi_error2}
\R(h_{\mathcal{H}_{\sigma, n}}) - R(h^{*}) \leq \dfrac{ \mathcal{K}_{\ell}}{ \big( n^{(\alpha)} \big)^{\frac{s}{s + d}(1-\frac{1}{r}) } }.
\end{equation}

\medskip

\noindent
\medskip

\noindent
\textbf{Step 3 : Expected excess risk bound}.
\medskip

\noindent
According to (\ref{Exp_excess_risk_v2}), (\ref{equa_bound_stochastic_error_v2}) and (\ref{approxi_error2}), we have,
\begin{align*}
\E[R(\widehat{h}_n) - R(h^{*})]  
& \leq \dfrac{ \big(\log n^{(\alpha)}  \big)^{\nu}}{ \big( n^{(\alpha)} \big)^{\frac{s}{s + d}(1-\frac{1}{r}) } }   + \frac{64\mathcal{K}_{\ell}  (1 + 4e^{-2}\overline{\alpha})}{3 \big(n^{(\alpha)} \big)^{(1 - \frac{1}{r})} }   +  \dfrac{6\mathcal{K}_{\ell} M}{ (n^{(\alpha)})^{(1 - \frac{1}{r})}} + \frac{3\mathcal{K}_{\ell}}{n^{(\alpha)}} + \dfrac{ \mathcal{K}_{\ell}}{ \big( n^{(\alpha)} \big)^{\frac{s}{s + d}(1-\frac{1}{r}) } } \\
& \leq \dfrac{ \big(\log n^{(\alpha)}  \big)^{\nu} + \mathcal{K}_{\ell} }{ \big( n^{(\alpha)} \big)^{\frac{s}{s + d}(1-\frac{1}{r}) } }   + \frac{ C(\mathcal{K}_{\ell}, \overline{\alpha}, M)  }{ \big(n^{(\alpha)} \big)^{(1 - \frac{1}{r})} }   + \frac{3\mathcal{K}_{\ell}}{n^{(\alpha)}},
\end{align*}
with $C(\mathcal{K}_{\ell}, \overline{\alpha}, M) =   64\mathcal{K}_{\ell}  (1 + 4e^{-2}\overline{\alpha})/3  +  6\mathcal{K}_{\ell} M $.
%
%
\qed

\subsection{Proof of Theorem  \ref{thm2}}
Let $ D_n = \{(X_i, Y_i)_{i = 1}^n)\}$ and $ D_n' = \{(X_i', Y_i')_{i = 1}^n) \}$ be two independent sets of observations generated, respectively, from a stationary and ergodic process
$ Z = \mathcal{X} \times \mathcal{Y}$ and $ Z' = \mathcal{X'} \times \mathcal{Y'}$, where $\mathcal{X}, \mathcal{X'} \subset \R^d$, and $\mathcal{Y}, \mathcal{Y'} \subset \R$.

\medskip

\noindent
Let $L_n, N_n, B_n, F_n, S_n > 0$  fulfill the conditions in Theorem \ref{thm2}. In the sequel, we set:
\begin{equation*}
\mathcal{H}_{\sigma, n} \coloneqq \mathcal{H}_{\sigma}(L_n, N_n, B_n, F_n, S_n).
\end{equation*}
Recall the inequality (\ref{Exp_excess_risk_v2}) (see proof of Theorem \ref{thm1}):
\begin{align}\label{Exp_excess_risk_thm2}
\E[R(\widehat{h}_n) - R(h^{*})] 
\leq  \E_{D_n} \Big[\dfrac{1}{n}\sum_{i= 1}^n \{ -2g(\widehat{h}_n, Z_i)+ \E_{D_n'} g(\widehat{h}_n, Z_i') \} \Big] + 2 [R(h_{\mathcal{H}_{\sigma, n}}) - R(h^{*})], 
\end{align}
where $g(h,Z)$ is defined in (\ref{proof_def_g}).
We will focus on the bounds of $\E_{D_n} \Big[\dfrac{1}{n}\sum_{i= 1}^n \{ -2g(\widehat{h}_n, Z_i)+ \E_{D_n'} g(\widehat{h}_n, Z_i') \} $ and $[R(h_{\mathcal{H}_{\sigma, n}}) - R(h^{*})]$.

\medskip

\noindent
For all $n \ge 1$, set
\begin{equation}\label{proof_def_beta_n_thm2}
\beta_n =  n^{\frac{\mu + 1}{r(2\mu + 3)} },
\end{equation}
for $\mu > 0, r >1$ given in the assumptions \textbf{(A3)} and \textbf{(A5)} respectively. 

\medskip

\noindent
\textbf{Step 1 : Bounding the first term in the right-hand side of (\ref{Exp_excess_risk_thm2})}.

\medskip

\noindent
Recall the inequality (\ref{proof_exp_2_g_ing}) (see proof of Theorem \ref{thm1}),
\begin{align}\label{proof_exp_2_g_ing_thm2} 
\E_{D_n} \Big[\dfrac{1}{n}\sum_{i= 1}^n \{ -2g(\widehat{h}_n, Z_i)+ \E_{D_n'} g(\widehat{h}_n, Z_i') \} \Big] \leq \E_{D_n}\Big[ \dfrac{1}{n} \sum_{i = 1}^n G_{\beta_n}(h_{j^*}, Z_i) \Big] +  6\mathcal{K}_{\ell} \E[|Y_i|^r]/ \beta_n^{r-1} + 3\mathcal{K}_{\ell}\epsilon,     
\end{align} 
where $G_{\beta_n}(h, Z)$ is defined in (\ref{def_g_beta}).
Let $ \epsilon >0$. 
Recall that, the $\epsilon$-covering number of $\mathcal{H}_{\sigma, n}$ is finite, the notation:
$ m \coloneqq \mathcal{N}\Big( \mathcal{H}_{\sigma, n}, \epsilon  \Big)$ and that
$\mathcal{H}_{\sigma,n} \subset \bigcup_{j=1}^m B(h_j, \epsilon)$, 
for $h_1,\cdots,h_m \in \mathcal{H}_{\sigma,n}$.
Let us derive an upper bound of $\E_{D_n}\Big[ \dfrac{1}{n} \sum_{i = 1}^n G_{\beta_n}(h_{j^*}, Z_i) \Big]$.
Let $ h \in \mathcal{H}_{\sigma, n}$. Since $F_n \leq \beta_n$, recall the inequality (\ref{proof_g_beta_K_ell_beta}):
$ |g_{\beta_n}(h, Z_i)|  \leq \mathcal{K}_{\ell}(F_n +\beta_n) \leq 2\mathcal{K}_{\ell}\beta_n$.    
%
%
\newline
Let $ \varepsilon > 0 $. 
From \cite{doukhan2007probability} and the Remark 3.3 in \cite{diop2022statistical},  we have for all $j=1,\cdots,m$,
\begin{align}
 \nonumber P \Bigg\{\dfrac{1}{n} \sum_{i=1}^n G_{\beta_n}(h_{j}, Z_i) > \varepsilon \Bigg\}   & = P \Bigg\{\dfrac{1}{n} \sum_{i=1}^n\E_{D_n'}[g_{\beta_n}(h_{j}, Z_i')] - \dfrac{2}{n} \sum_{i = 1}^n g_{\beta_n}(h_{j}, Z_i) > \varepsilon \Bigg\}
\\
\nonumber & =  P \Big\{\E_{D_n'}[g_{\beta_n}(h_{j}, Z_1')] - \dfrac{2}{n} \sum_{i = 1}^n g_{\beta_n}(h_{j}, Z_i) > \varepsilon \Big\}.
\\
\nonumber & =  P \Big\{2\E_{D_n'}[g_{\beta_n}(h_{j}, Z_1')] - \dfrac{2}{n} \sum_{i = 1}^n g_{\beta_n}(h_j, Z_i) > \varepsilon + \E_{D_n'}[g_{\beta_n}(h_{j}, Z_1')] \Big\}
\\
\nonumber & =  P \Big\{\E_{D_n'}[g_{\beta_n}(h_{j}, Z_1')] - \dfrac{1}{n} \sum_{i = 1}^n g_{\beta_n}(h_{j}, Z_i) > \dfrac{\varepsilon}{2} + \dfrac{1}{2} \E_{D_n'}[g_{\beta_n}(h_{j^*}, Z_1')]\Big\}
\\
\nonumber & \leq  P \Bigg\{\E_{D_n'}[g_{\beta_n}(h_{j}, Z_1')] - \dfrac{1}{n} \sum_{i = 1}^n g_{\beta_n}(h_{j}, Z_i) > \dfrac{\varepsilon}{2} + \dfrac{\E_{D_n'}[|g_{\beta_n}(h_{j}, Z_1')|^2]}{4 \mathcal{K}_{\ell}\beta_n} \Bigg\}
\\
\nonumber & \leq  P \Big\{ \E[g_{\beta_n}(h_{j}, Z_1)] - \dfrac{1}{n} \sum_{i = 1}^n g_{\beta_n}(h_{j}, Z_i) > \dfrac{\varepsilon}{2} \Big\} \\
& \leq \exp\Bigg(- \frac{n^{2}\varepsilon^2/16}{n C_{n, 1} + 2C_{n, 2}^{1/(\mu+2)} \big( n\varepsilon/2\big)^{(2\mu+3)/(\mu+2)}} \Bigg),
\end{align}
where $ C_{n, 1} =  16\mathcal{K}_{\ell}^2\beta_n^2\Psi(1, 1)L_{1}, C_{n, 2} = 4\mathcal{K}_{\ell}\beta_n L_{2}\max(2^{3 + \mu}/\Psi (1, 1), 1)$. 

\medskip

\noindent
Take $\epsilon = \frac{1}{n}$, that is $m = \mathcal{N}\Big( \mathcal{H}_{\sigma, n},  1/n \Big)$. In addition to (\ref{proof_ing_covering_num}) we have,
\begin{align}\label{proof_bound_prob_v1}
\nonumber P \Big\{\dfrac{1}{n} \sum_{i = 1}^n G_{\beta_n}(h_{j^*}, Z_i) > \varepsilon \Big\}  & \leq   P \Bigg\{ \bigcup_{j=1}^m \Big( \dfrac{1}{n} \sum_{i = 1}^n G_{\beta_n}(h_{j}, Z_i) > \varepsilon \Big) \Bigg\} \leq \sum_{j=1}^m  P \Big\{\dfrac{1}{n} \sum_{i = 1}^n G_{\beta_n}(h_{j}, Z_i) > \varepsilon \Big\}   \\
\nonumber & \leq \mathcal{N}\Big( \mathcal{H}_{\sigma, n}, \frac{1}{n}\Big)\cdot \exp\Bigg(- \frac{n^{2}\varepsilon^2/16}{n C_{n, 1} + 2C_{n, 2}^{1/(\mu+2)} \big( n\varepsilon/2\big)^{(2\mu+3)/(\mu+2)}} \Bigg)
\\
& \leq \exp \Bigg(2L_n(S_n + 1)\log\Big(n C_{\sigma}L_n(N_n + 1)(B_n \lor 1)\Big) - \frac{n^{2}\varepsilon^2/16}{n C_{n, 1} + 2C_{n, 2}^{1/(\mu+2)} \big( n\varepsilon/2\big)^{(2\mu+3)/(\mu+2)}} \Bigg). 
\end{align}
Moreover,
\begin{align}\label{proof_value_epsi_n}
2C_{n, 2}^{1/(\mu+2)} \big( n\varepsilon/2\big)^{(2\mu+3)/(\mu+2)} & > n C_{n, 1} 
 \Longrightarrow \varepsilon  > \dfrac{2^{(\mu + 1)/(2\mu + 3) } C_{n, 1}^{(\mu + 2)/(2\mu + 3)}}{n^{(\mu + 1)/(2\mu + 3)} C_{n, 2}^{1/(2\mu + 3)}} \coloneqq \varepsilon_n.
 \end{align}
One can see that, $ \dfrac{ C_{n, 1}^{(\mu + 2)/(2\mu + 3)}}{C_{n, 2}^{1/(2\mu + 3)}} = \dfrac{\big(16\mathcal{K}_{\ell}^2\Psi(1, 1)L_{1} \big)^{\frac{\mu + 2}{2\mu + 3}} \beta_n}{\big( 4\mathcal{K}_{\ell} L_{2}\max(2^{3 + \mu}/\Psi (1, 1), 1)\big)^{\frac{1}{2\mu + 3}}} $. 

\medskip

\noindent
Thus, we have,
$ \varepsilon_n \coloneqq \dfrac{2^{(\mu + 1)/(2\mu + 3) }\big(16\mathcal{K}_{\ell}^2\Psi(1, 1)L_{1} \big)^{\frac{\mu + 2}{2\mu + 3}} }{\big( 4\mathcal{K}_{\ell} L_{2}\max(2^{3 + \mu}/\Psi (1, 1), 1)\big)^{\frac{1}{2\mu + 3}}n^{ (1 - 1/r )\frac{\mu + 1}{2\mu + 3}}} $.

\medskip

\noindent
From (\ref{proof_value_epsi_n}), we get,
\begin{align}\label{proof_cond_denom_v1} 
-\dfrac{1}{n C_{n, 1} + 2C_{n, 2}^{1/(\mu+2)} \big(n\varepsilon/2 \big)^{(2\mu+3)/(\mu+2)}} \leq - \dfrac{1}{ 4C_{n, 2}^{1/(\mu+2)} \big(n\varepsilon/2 \big)^{(2\mu+3)/(\mu+2)} }. 
\end{align}
According to (\ref{proof_bound_prob_v1}) and (\ref{proof_cond_denom_v1}), we have,
\begin{align}\label{proof_bound_proba_v2}
  P \Big\{\dfrac{1}{n} \sum_{i = 1}^n G_{\beta_n}(h_i, Z_i)  > \varepsilon \Big\} \leq \exp\Bigg(2L_n(S_n + 1) \log\Big(n C_{\sigma}L_n(N_n + 1)(B_n\lor 1)\Big) 
 - \frac{n^{1/{(\mu + 2)}} \varepsilon^{1/(\mu + 2)}}{64 C_{n, 2}^{1/(\mu + 2)} (1/2)^{(2\mu+3)/(\mu+2)}} \Bigg).
\end{align}
\textbf{Case 1}: $\varepsilon > 1 > \varepsilon_n$.

\medskip

\noindent
Recall that from (\ref{proof_bound_proba_v2}), we have,
\begin{align}\label{equa_bound_stochas_error_v1}
P \Big\{\dfrac{1}{n} \sum_{i = 1}^n G_{\beta_n}(h_i, Z_i)  > \varepsilon \Big\} 
& \leq \exp\Bigg(2L_n(S_n + 1) \log\Big(n C_{\sigma}L_n(N_n + 1)(B_n\lor 1)\Big) 
 - \frac{n^{1/{(\mu + 2)}} \varepsilon^{1/(\mu + 2)}}{64 C_{n, 2}^{1/(\mu + 2)} (1/2)^{(2\mu+3)/(\mu+2)}} \Bigg).
\end{align}
Recall the assumptions:
\begin{equation}\label{equa_networks_archi}
L_n = \big(1 - \frac{1}{r} \big) \dfrac{s L_0}{s + d} \log(n ), N_n = N_0 n^{\big(1 - \frac{1}{r} \big) \frac{d}{s + d}}, S_n = \big(1 - \frac{1}{r} \big) \frac{s S_0}{s + d} n^{\big(1 - \frac{1}{r} \big) \frac{d}{s + d}}\log(n),  B_n = B_0 n^{ \big(1 - \frac{1}{r} \big) \frac{4s(d/s + 1)}{s + d}},
\end{equation}
with $L_0, N_0, B_0, S_0 >0$. In addition to (\ref{equa_networks_archi}), we have,
\begin{align*}
& \nonumber P \Big\{\dfrac{1}{n} \sum_{i = 1}^n G_{\beta_n}(h_i, Z_i) > \varepsilon \Big\}  \\
\nonumber & \leq \exp\Bigg(2\big(1 - \frac{1}{r} \big) \dfrac{s L_0}{s + d} \log(n) \big(\big(1 - \frac{1}{r} \big) \frac{s S_0}{s + d} n^{\big(1 - \frac{1}{r} \big) \frac{d}{s + d}}\log(n) + 1 \big) \\
 & \times \log\Big(n C_{\sigma} \big(1 - \frac{1}{r} \big) \dfrac{s L_0}{s + d} \log(n) \big(N_0 n^{\big(1 - \frac{1}{r} \big) \frac{d}{s + d}} + 1 \big)(B_0 n^{ \big(1 - \frac{1}{r} \big) \frac{4s(d/s + 1)}{s + d}}\lor 1) \Big) 
 - \frac{n^{1/{(\mu + 2)}} \varepsilon^{1/(\mu + 2)}}{64 C_{n, 2}^{1/(\mu + 2)} (1/2)^{(2\mu+3)/(\mu+2)}} \Bigg).
\end{align*}
Thus, we have
\begin{align*}
& \nonumber P \Big\{\dfrac{1}{n} \sum_{i = 1}^n G_{\beta_n}(h_i, Z_i) > \varepsilon \Big\}  \\
\nonumber & \leq \exp\Bigg[ \frac{n^{1/{(\mu + 2)}} \varepsilon^{1/(\mu + 2)}}{64 C_{n, 2}^{1/(\mu + 2)} (1/2)^{(2\mu+3)/(\mu+2)}}  \\
& \hspace{3.4cm}  \times \Bigg( \dfrac{128 C_{n, 2}^{1/(\mu + 2)} (1/2)^{(2\mu+3)/(\mu+2)}\big(1 - \frac{1}{r} \big) \dfrac{s L_0}{s + d} \log(n)  \big(\big(1 - \frac{1}{r} \big) \frac{s S_0}{s + d} n^{\big(1 - \frac{1}{r} \big) \frac{d}{s + d}}\log(n) + 1 \big)  }{n^{1/{(\mu + 2)}}} \\
 & \hspace{5cm} \times \log\Big(n C_{\sigma} \big(1 - \frac{1}{r} \big) \dfrac{s L_0}{s + d} \log(n) \big(N_0 n^{\big(1 - \frac{1}{r} \big) \frac{d}{s + d}} + 1 \big)(B_0 n^{ \big(1 - \frac{1}{r} \big) \frac{4s(d/s + 1)}{s + d}}\lor 1) \Big) 
 - 1 \Bigg) \Bigg].
\end{align*}
According to the condition $ s > d \Big( \dfrac{(r-1)(2\mu+3)(\mu +2)}{ r(2\mu +3) - (\mu+1)} - 1 \Big)$, it holds that,
\begin{align*}
 & \nonumber \dfrac{128 C_{n, 2}^{1/(\mu + 2)} (1/2)^{(2\mu+3)/(\mu+2)}\big(1 - \frac{1}{r} \big) \dfrac{s L_0}{s + d} \log(n) \big(\big(1 - \frac{1}{r} \big) \frac{s S_0}{s + d} n^{\big(1 - \frac{1}{r} \big) \frac{d}{s + d}}\log(n) + 1 \big)}{n^{1/{(\mu + 2)}}} \\
& \times \log\Big(n C_{\sigma} \big(1 - \frac{1}{r} \big) \dfrac{s L_0}{s + d} \log(n) \big(N_0 n^{\big(1 - \frac{1}{r} \big) \frac{d}{s + d}} + 1 \big)(B_0 n^{ \big(1 - \frac{1}{r} \big) \frac{4s(d/s + 1)}{s + d}}\lor 1) \Big) \limiten 0.   
\end{align*}
Hence, there exists $n_1 = n_1(L_0, N_0, B_0, S_0, C_{\sigma}, r, \mu, s, d)$ such that, for any $n \ge n_1$, we have,
\begin{align}\label{equa_limite_thm2}
 & \nonumber \dfrac{128 C_{n, 2}^{1/(\mu + 2)} (1/2)^{(2\mu+3)/(\mu+2)}\big(1 - \frac{1}{r} \big) \dfrac{s L_0}{s + d} \log(n) \big(\big(1 - \frac{1}{r} \big) \frac{s S_0}{s + d} n^{\big(1 - \frac{1}{r} \big) \frac{d}{s + d}}\log(n) + 1 \big)}{n^{1/{(\mu + 2)}}} \\
& \times \log\Big(n C_{\sigma} \big(1 - \frac{1}{r} \big) \dfrac{s L_0}{s + d} \log(n) \big(N_0 n^{\big(1 - \frac{1}{r} \big) \frac{d}{s + d}} + 1 \big)(B_0 n^{ \big(1 - \frac{1}{r} \big) \frac{4s(d/s + 1)}{s + d}}\lor 1) \Big) < \dfrac{1}{2}.   
\end{align}
According to (\ref{equa_limite_thm2}), we have
\begin{equation*}
 P \Big\{\dfrac{1}{n} \sum_{i = 1}^n G_{\beta_n}(h_i, Z_i) > \varepsilon \Big\}   \leq \exp\Big( -\frac{n^{1/{(\mu + 2)}} \varepsilon^{1/(\mu + 2)}}{128 C_{n, 2}^{1/(\mu + 2)} (1/2)^{(2\mu+3)/(\mu+2)}} \Big),
\end{equation*}
which holds for all  $\varepsilon > 1$.
Hence,
\begin{align}\label{equa_bound_proba_thm2_v1}
\displaystyle \int_{1}^{\infty} P \Big\{\dfrac{1}{n} \sum_{i = 1}^n G_{\beta_n}(h_i, Z_i)  > \varepsilon \Big\} d\varepsilon & \leq \displaystyle \int_{1}^{\infty} \exp\Big( -\frac{n^{1/{(\mu + 2)}} \varepsilon^{1/(\mu + 2)}}{128 C_{n, 2}^{1/(\mu + 2)} (1/2)^{(2\mu+3)/(\mu+2)}} \Big) d\varepsilon.
\end{align}
To compute the right-hand side in (\ref{equa_bound_proba_thm2_v1}), we use similar arguments as in \cite{kengne2023penalized}.

\medskip

Let
\[ \widetilde{x} = \dfrac{n^{1/(\mu + 2)} }{ 128 C_{n, 2}^{1/(\mu + 2)} (1/2)^{(2\mu+3)/(\mu+2)} } \varepsilon^{1/(\mu + 2)},  ~ g(x) \coloneqq x - 2(\mu + 2) \log(x). \]
Let $ x_0 = (2(\mu + 2))^3 $.
One can  easily see  that  $ g(x_0) > 0$ and to get
\[ \widetilde{x} > x_0  ~ \text{that is} \quad \dfrac{n^{1/(\mu + 2)} }{ 128 C_{n, 2}^{1/(\mu + 2)} (1/2)^{(2\mu+3)/(\mu+2)} } \varepsilon^{1/(\mu + 2)} >  (2(\mu + 2))^3, \quad \forall \varepsilon > 1,  \]
it suffice that,
\[
n > \Big( 128 C(\mathcal{K}_\ell, \mu)^{1/(\mu + 2)} (1/2)^{(2\mu+3)/(\mu+2)} (2(\mu + 2))^3 \Big)^{ \frac{r(2\mu+3)(\mu+2)}{ r(2\mu+3) - (\mu+1)}},
\]
with $ C(\mathcal{K}_\ell, \mu) = 4\mathcal{K}_{\ell} L_{2}\max(2^{3 + \mu}/\Psi (1, 1), 1)$.
One can remark that for any 
\begin{equation*}
x > x_0, \quad \text{we have} \quad  \exp(- x) <  \dfrac{1}{x^{2(\mu + 2)}}.
\end{equation*}
Therefore, (\ref{equa_bound_proba_thm2_v1}) gives,
\begin{align*}
\nonumber \displaystyle \int_{1}^{\infty} \exp\Big( -\dfrac{n^{1/{(\mu + 2)}} \varepsilon^{1/(\mu + 2)}}{128 C_{n, 2}^{1/(\mu + 2)} (1/2)^{(2\mu+3)/(\mu+2)}} \Big) d\varepsilon & \leq \displaystyle \int_{1}^{\infty} \dfrac{1}{\Big(\dfrac{n^{1/(\mu + 2)} }{128 C_{n, 2}^{1/(\mu + 2)} (1/2)^{(2\mu+3)/(\mu+2)}} \varepsilon^{1/(\mu + 2)}\Big)^{2(\mu + 2)}} d\varepsilon  \\
& \leq \dfrac{(16384)^{\mu + 2}(1/2)^{4\mu+6} C_{2, n}^2}{n^2}\displaystyle \int_{1}^{\infty} \dfrac{1}{\varepsilon^2} d\varepsilon = \dfrac{(16384)^{\mu + 2}(1/2)^{4\mu+6} C_{2, n}^2}{n^2}. 
\end{align*}
Thus, we have
\begin{align*}
\displaystyle \int_{1}^{\infty} P \Big\{\dfrac{1}{n} \sum_{i = 1}^n G_{\beta_n}(h_i, Z_i)  > \varepsilon \Big\} d\varepsilon \leq
 \dfrac{(16384)^{\mu + 2}(1/2)^{4\mu+6} C_{2, n}^2}{n^2}.  
\end{align*}
Recall that we have set $C_{2, n} = 4\mathcal{K}_{\ell}\beta_n L_{2}\max(2^{3 + \mu}/\Psi (1, 1), 1)$. 
Since $\beta_n = n^{\frac{\mu+1}{r(2\mu + 3)}}$ (see (\ref{proof_def_beta_n_thm2})),  we have,
\begin{equation}\label{equa_bound_case1}
\displaystyle \int_{1}^{\infty} P \Big\{\dfrac{1}{n} \sum_{i = 1}^n G_{\beta_n}(h_i, Z_i)  > \varepsilon \Big\} d\varepsilon \leq
 \dfrac{(16384)^{\mu + 2}(1/2)^{4\mu+6} C(\mathcal{K}_\ell, \mu)^2}{n^{2\big(1 - \frac{\mu+1}{r(2\mu + 3)} \big)}},  
\end{equation}
where $C(\mathcal{K}_\ell, \mu) = 4\mathcal{K}_{\ell} L_{2}\max(2^{3 + \mu}/\Psi (1, 1), 1)$. \\
So, one can find  $n_2 = n_2(\mu, \mathcal{K}_{\ell})$ such that, for all $n > n_2$, we have,
\begin{equation}\label{proof_bound_prob_case1}
 \int_{1}^{\infty} P \Big\{\dfrac{1}{n} \sum_{i = 1}^n G_{\beta_n}(h_i, Z_i)  > \varepsilon \Big\} d\varepsilon \leq
 \dfrac{(16384)^{\mu + 2}(1/2)^{4\mu+6} C(\mathcal{K}_\ell, \mu)^2}{n^{2\big(1 - \frac{\mu+1}{r(2\mu + 3)} \big)}} \leq \dfrac{1}{n^{\big(1 - 1/r \big)(\mu + 1)/(2\mu + 3)}}.
\end{equation}

\medskip

\noindent
\textbf{Case 2}: $\varepsilon_n < \varepsilon < 1$.
Recall that from (\ref{proof_bound_proba_v2}), we have,
\begin{align*}
P \Big\{\dfrac{1}{n} \sum_{i = 1}^n G_{\beta_n}(h_i, Z_i)  > \varepsilon \Big\} 
& \leq \exp\Bigg(2L_n(S_n + 1) \log\Big(n C_{\sigma}L_n(N_n + 1)(B_n\lor 1)\Big) 
 - \frac{n^{1/{(\mu + 2)}} \varepsilon^{1/(\mu + 2)}}{64 C_{n, 2}^{1/(\mu + 2)} (1/2)^{(2\mu+3)/(\mu+2)}} \Bigg) \\
& \leq \exp\Bigg(2L_n(S_n + 1) \log\Big(n C_{\sigma}L_n(N_n + 1)(B_n\lor 1)\Big)
 - \frac{n^{1/{(\mu + 2)}} \varepsilon_n^{1/(\mu + 2)}}{64 C_{n, 2}^{1/(\mu + 2)} (1/2)^{(2\mu+3)/(\mu+2)} } \Bigg).
\end{align*}
According to (\ref{proof_value_epsi_n}) and (\ref{equa_networks_archi}), we have,
\begin{align*}
& \nonumber P \Big\{\dfrac{1}{n} \sum_{i = 1}^n G_{\beta_n}(h_i, Z_i) > \varepsilon \Big\}  \\
\nonumber & \leq \exp\Bigg(2\big(1 - \frac{1}{r} \big) \dfrac{s L_0}{s + d} \log(n) \big(\big(1 - \frac{1}{r} \big) \frac{s S_0}{s + d} n^{\big(1 - \frac{1}{r} \big) \frac{d}{s + d}}\log(n) + 1 \big) \\
 &  \times \log\Big(n C_{\sigma} \big(1 - \frac{1}{r} \big) \dfrac{s L_0}{s + d} \log(n) \big(N_0 n^{\big(1 - \frac{1}{r} \big) \frac{d}{s + d}} + 1 \big)(B_0 n^{ \big(1 - \frac{1}{r} \big) \frac{4s(d/s + 1)}{s + d}}\lor 1) \Big) 
 - \frac{n^{1/{(2\mu + 3)}} 2^{(\mu + 1)/(2\mu + 3)(\mu + 2) } C_{n, 1}^{1/(2\mu + 3)}}{64 C_{n, 2}^{2/(2\mu + 3)} (1/2)^{(2\mu+3)/(\mu+2)} } \Bigg) \\
\nonumber & \leq \exp\Bigg( \frac{n^{1/{(2\mu + 3)}} 2^{(\mu + 1)/(2\mu + 3)(\mu + 2) } C_{n, 1}^{1/(2\mu + 3)}}{64 C_{n, 2}^{2/(2\mu + 3)} (1/2)^{(2\mu+3)/(\mu+2)}} \\
& \hspace{4cm} \times \Bigg( \dfrac{128 C_{n, 2}^{2/(2\mu + 3)} (1/2)^{(2\mu+3)/(\mu+2)} \big(1 - \frac{1}{r} \big) \dfrac{s L_0}{s + d} \log(n) \big(\big(1 - \frac{1}{r} \big) \frac{s S_0}{s + d} n^{\big(1 - \frac{1}{r} \big) \frac{d}{s + d}}\log(n) + 1 \big)}{n^{1/{(2\mu + 3)}} 2^{(\mu + 1)/(2\mu + 3)(\mu + 2) } C_{n, 1}^{1/(2\mu + 3)}} \\
 & \hspace{4cm} \times \log\Big(n C_{\sigma} \big(1 - \frac{1}{r} \big) \dfrac{s L_0}{s + d} \log(n) \big(N_0 n^{\big(1 - \frac{1}{r} \big) \frac{d}{s + d}} + 1 \big)(B_0 n^{ \big(1 - \frac{1}{r} \big) \frac{4s(d/s + 1)}{s + d}}\lor 1) \Big) 
 - 1 \Bigg) \Bigg).
\end{align*}

\medskip

\noindent
Recall the condition $ s > d \Big( \big(1 - \frac{1}{r} \big)(2\mu +3) -1 \Big)$, that is $\dfrac{1}{2\mu + 3} > \big(1 - \dfrac{1}{r} \big)\dfrac{d}{s + d}$. 
Consequently,
\begin{align*}
& \nonumber \dfrac{128 C_{n, 2}^{2/(2\mu + 3)} (1/2)^{(2\mu+3)/(\mu+2)}\big(1 - \frac{1}{r} \big) \dfrac{s L_0}{s + d} \log(n) \big(\big(1 - \frac{1}{r} \big) \frac{s S_0}{s + d} n^{\big(1 - \frac{1}{r} \big) \frac{d}{s + d}}\log(n) + 1 \big)}{n^{1/{(2\mu + 3)}} 2^{(\mu + 1)/(2\mu + 3)(\mu + 2) } C_{n, 1}^{1/(2\mu + 3)}} \\
 &  \times \log\Big(n C_{\sigma} \big(1 - \frac{1}{r} \big) \dfrac{s L_0}{s + d} \log(n) \big(N_0 n^{\big(1 - \frac{1}{r} \big) \frac{d}{s + d}} + 1 \big)(B_0 n^{ \big(1 - \frac{1}{r} \big) \frac{4s(d/s + 1)}{s + d}}\lor 1) \Big) \limiten 0.   
\end{align*}
Then, there exists $n_3 = n_3(L_0, N_0, B_0, S_0, C_{\sigma}, r, L_1, L_2, \mu, s, d, r)$ such that, for any $n > n_3$, we have,
\begin{align}\label{equa_bound_limite_v2}
& \nonumber \dfrac{128 C_{n, 2}^{2/(2\mu + 3)} (1/2)^{(2\mu+3)/(\mu+2)}\big(1 - \frac{1}{r} \big) \dfrac{s L_0}{s + d} \log(n) \big(\big(1 - \frac{1}{r} \big) \frac{s S_0}{s + d} n^{\big(1 - \frac{1}{r} \big) \frac{d}{s + d}}\log(n) + 1 \big)}{n^{1/{(2\mu + 3)}} 2^{(\mu + 1)/(2\mu + 3)(\mu + 2) } C_{n, 1}^{1/(2\mu + 3)}} \\
 &  \times \log\Big(n C_{\sigma} \big(1 - \frac{1}{r} \big) \dfrac{s L_0}{s + d} \log(n) \big(N_0 n^{\big(1 - \frac{1}{r} \big) \frac{d}{s + d}} + 1 \big)(B_0 n^{ \big(1 - \frac{1}{r} \big) \frac{4s(d/s + 1)}{s + d}}\lor 1) \Big) < \dfrac{1}{2}.   
\end{align}
According to (\ref{equa_bound_limite_v2}), we have
\begin{align*}
P \Big\{\dfrac{1}{n} \sum_{i = 1}^n G_{\beta_n}(h_i, Z_i) > \varepsilon \Big\} \leq \exp\Bigg( -\frac{n^{1/{(2\mu + 3)}} 2^{(\mu + 1)/(2\mu + 3)(\mu + 2) } C_{n, 1}^{1/(2\mu + 3)}}{128 C_{n, 2}^{2/(2\mu + 3)} (1/2)^{(2\mu+3)/(\mu+2)}}  \Bigg).
\end{align*}
Hence,
\begin{align}\label{equa_bound_case2}
\nonumber \displaystyle \int_{\varepsilon_n}^1 P \Big\{\dfrac{1}{n} \sum_{i = 1}^n G_{\beta_n}(h_i, Z_i)  > \varepsilon \Big\} 
& \leq \displaystyle \int_{\varepsilon_n}^1  \exp\Bigg( -\frac{n^{1/{(2\mu + 3)}} 2^{(\mu + 1)/(2\mu + 3)(\mu + 2) } C_{n, 1}^{1/(2\mu + 3)}}{128 C_{n, 2}^{2/(2\mu + 3)} (1/2)^{(2\mu+3)/(\mu+2)}}  \Bigg) d\varepsilon \\
\nonumber & \leq (1 - \varepsilon_n)\exp\Bigg( -\frac{n^{1/{(2\mu + 3)}} 2^{(\mu + 1)/(2\mu + 3)(\mu + 2) } C_{n, 1}^{1/(2\mu + 3)}}{128 C_{n, 2}^{2/(2\mu + 3)} (1/2)^{(2\mu+3)/(\mu+2)}}  \Bigg) \\
\nonumber & \leq \Big(1 -  \dfrac{2^{(\mu + 1)/(2\mu + 3) } C_{n, 1}^{(\mu + 2)/(2\mu + 3)}}{n^{(\mu + 1)/(2\mu + 3)} C_{n, 2}^{1/(2\mu + 3)}} \Big)\exp\Bigg( -\frac{n^{1/{(2\mu + 3)}} 2^{(\mu + 1)/(2\mu + 3)(\mu + 2) } C_{n, 1}^{1/(2\mu + 3)}}{128 C_{n, 2}^{2/(2\mu + 3)} (1/2)^{(2\mu+3)/(\mu+2)}}  \Bigg)\\
 \nonumber & \leq \exp\Bigg( -\frac{n^{1/{(2\mu + 3)}} 2^{(\mu + 1)/(2\mu + 3)(\mu + 2) } C_{n, 1}^{1/(2\mu + 3)}}{128 C_{n, 2}^{2/(2\mu + 3)} (1/2)^{(2\mu+3)/(\mu+2)}}  \Bigg) \\
&\leq \exp\Bigg( -\frac{n^{1/{(2\mu + 3)}} 2^{(\mu + 1)/(2\mu + 3)(\mu + 2)} C(\mu, L_1, L_2)}{128(1/2)^{(2\mu+3)/(\mu+2)}}  \Bigg), 
\end{align}
with $C(\mu, L_1, L_2) := \dfrac{C_{n, 1}^{1/(2\mu + 3)} }{C_{n, 2}^{2/(2\mu + 3)}}  = \dfrac{ \big( \Psi(1, 1)L_1 \big)^{1/(2\mu +3)}}{\Big(L_2 \max \big(2^{3 + \mu})/\Psi(1, 1), 1 \big)\Big)^{2/(2\mu + 3)}}$. \\
One can find  $n_4 = n_4(\mu, L_1, L_2)$ such that, for all $n > n_4$, we have,
\begin{equation}\label{proof_bound_prob_case2}
\int_{\varepsilon_n}^1 P \Big\{\dfrac{1}{n} \sum_{i = 1}^n G_{\beta_n}(h_i, Z_i)  > \varepsilon \Big\} \leq \exp\Bigg( -\frac{n^{1/{(2\mu + 3)}} 2^{(\mu + 1)/(2\mu + 3)(\mu + 2)} C(\mu, L_1, L_2)}{128(1/2)^{(2\mu+3)/(\mu+2)}}  \Bigg) \leq \dfrac{1}{n^{\big(1 - 1/r \big)(\mu + 1)/(2\mu + 3)}}
\end{equation}

\medskip

\noindent
\textbf{Case 3}: $ 0 < \varepsilon < \varepsilon_n$.
We have 
\begin{align*}
\displaystyle \int_{0}^{\varepsilon_n}P \Big\{\dfrac{1}{n} \sum_{i = 1}^n G_{\beta_n}(h_i, Z_i)  > \varepsilon \Big\} d\varepsilon & \leq \displaystyle \int_{0}^{\varepsilon_n} d\varepsilon \leq \varepsilon_n = \dfrac{2^{(\mu + 1)/(2\mu + 3) } C_{n, 1}^{(\mu + 2)/(2\mu + 3)}}{n^{(\mu + 1)/(2\mu + 3)} C_{n, 2}^{1/(2\mu + 3)}}.
\end{align*}
We also get
\[ \dfrac{C_{n, 1}^{(\mu + 2)/(2\mu + 3)}}{C_{n, 2}^{1/(2\mu + 3)}} =  \dfrac{ 4\mathcal{K}_{\ell} \big(  \Psi(1, 1) L_1 \big)^{(\mu + 2)/(2\mu + 3)}}{\Big( L_2 \max \big( 2^{3 + \mu}/\Psi(1, 1), 1\big) \Big)^{1/(2\mu +3)}}  \beta_n \]
Set,
\begin{equation}\label{proof_def_C_mu_K_ell_L_1_L_2}
 C_0(\mu, \mathcal{K}_{\ell}, L_1, L_2) =  \dfrac{ 2^{(\mu + 1)/(2\mu + 3) } 4\mathcal{K}_{\ell} \big(  \Psi(1, 1) L_1 \big)^{(\mu + 2)/(2\mu + 3)}}{\Big( L_2 \max \big( 2^{3 + \mu}/\Psi(1, 1), 1\big) \Big)^{1/(2\mu +3)}} . 
\end{equation}
Since, $\beta_n = n^{\frac{\mu + 1}{r(2\mu + 3)}}$, we have,
\begin{align}\label{equa_bound_case3}
\displaystyle \int_{0}^{\varepsilon_n}P \Big\{\dfrac{1}{n} \sum_{i = 1}^n G_{\beta_n}(h_i, Z_i)  > \varepsilon \Big\} d\varepsilon \leq \dfrac{ C_0(\mu, \mathcal{K}_{\ell}, L_1, L_2)}{n^{\big(1 - 1/r \big)(\mu + 1)/(2\mu + 3)}}.
\end{align}

\medskip

Now, set
\[ n_0 \coloneqq \max\Bigg(n_1, n_2, n_3,n_4,   \Big( 128 C(\mathcal{K}_\ell, \mu)^{1/(\mu + 2)} (1/2)^{(2\mu+3)/(\mu+2)} (2(\mu + 2))^3 \Big)^{ \frac{r(2\mu+3)(\mu+2)}{ r(2\mu+3) - (\mu+1)}}  \Bigg) . \]
From, (\ref{proof_bound_prob_case1}), (\ref{proof_bound_prob_case2}) and (\ref{equa_bound_case3}), we have, for $n \geq n_0$,
\begin{align}\label{equa_bound_stochas_error_thm2}
\nonumber \E[\dfrac{1}{n} \sum_{i = 1}^n G_{\beta_n}(h_i, Z_i)] & \leq \displaystyle \int_{0}^{\infty}P \Big\{\dfrac{1}{n} \sum_{i = 1}^n G_{\beta_n}(h_i, Z_i)  > \varepsilon \Big\} d\varepsilon \\
  & \leq  \dfrac{C(\mu, \mathcal{K}_{\ell}, L_1, L_2)}{n^{\big(1 - 1/r \big)(\mu + 1)/(2\mu + 3)}} ,
\end{align}
with $  C(\mu, \mathcal{K}_{\ell}, L_1, L_2) := 2+ C_0(\mu, \mathcal{K}_{\ell}, L_1, L_2) $.
Thus, according (\ref{proof_exp_2_g_ing_thm2}) and (\ref{equa_bound_stochas_error_thm2}), we have, 
\begin{align*}
\E_{D_n} \Big[\dfrac{1}{n}\sum_{i= 1}^n \{ -2g(\widehat{h}_n, Z_i)+ \E_{D_n'} g(\widehat{h}_n, Z_i') \} \Big] \leq \dfrac{C(\mu, \mathcal{K}_{\ell}, L_1, L_2)}{n^{\big(1 - 1/r \big)(\mu + 1)/(2\mu + 3)}} +  6\mathcal{K}_{\ell} \dfrac{\E[|Y_i|^r]}{\beta_n^{r-1}} + 3\mathcal{K}_{\ell}\epsilon.     
\end{align*} 
Since, $\beta_n = n^{\frac{\mu+1}{2(2\mu + 3)}}, \E[|Y_i|^r \leq M$ and we have set $\epsilon= \dfrac{1}{n}$, it holds that,
\begin{align}\label{proof_exp_2_g_ing_thm2_v1} 
\E_{D_n} \Big[\dfrac{1}{n}\sum_{i= 1}^n \{ -2g(\widehat{h}_n, Z_i)+ \E_{D_n'} g(\widehat{h}_n, Z_i') \} \Big] \leq \dfrac{C(\mu, \mathcal{K}_{\ell}, L_1, L_2)}{n^{\big(1 - 1/r \big)(\mu + 1)/(2\mu + 3)}} +  \dfrac{6\mathcal{K}_{\ell} M}{n^{(1- 1/r)(\mu+1)/(2\mu+3)}} + \dfrac{3\mathcal{K}_{\ell}}{n}.     
\end{align} 

Let us derive a bound of the approximation error.

\medskip

\noindent
\textbf{Step 2: Bound the second term in the right-hand side of (\ref{Exp_excess_risk_thm2}).}

\medskip

\noindent
We set $\mathcal{H}_{\sigma, n} \coloneqq \mathcal{H}_{\sigma}(L_n, N_n, B_n, F_n, S_n)$ as in equation (\ref{proof_def_H_sigma_n}).
Recall the assumption $h^{*} \in C^{s, \mathcal{K}}(\mathcal{X})$ for some $s, \mathcal{K} > 0$.
Set $  \epsilon_n = \dfrac{ 1}{n^{\frac{s}{s + d}(1-\frac{1}{r}) } } $.
From \cite{kengne2023excess}, there exists a positive constants  $L_0, N_0, B_0, S_0 >0$ such that with $L_n = \big(1 - \frac{1}{r} \big) \dfrac{s L_0}{s + d} \log(n), N_n = N_0 n^{\big(1 - \frac{1}{r} \big) \frac{d}{s + d}}, S_n = \big(1 - \frac{1}{r} \big) \frac{s S_0}{s + d} n^{\big(1 - \frac{1}{r} \big) \frac{d}{s + d}}\log(n), 
$
$ B_n = B_0 n^{ \big(1 - \frac{1}{r} \big) \frac{4s(d/s + 1)}{s + d}}$, there is a neural network $h_{n} \in \mathcal{H}_{\sigma, n}$ satisfying,
\begin{equation}\label{cond_excess_risk_bound_thm2}
\| h_{n} - h^{*} \|_{\infty, \mathcal{X}} < \dfrac{ 1}{ n^{\frac{s}{s + d}(1-\frac{1}{r}) } }.
\end{equation}
According to (\ref{equa_approxi_error1}) and in addition to (\ref{cond_excess_risk_bound_thm2}), we have,
\begin{equation}\label{approxi_error2_thm2}
\R(h_{\mathcal{H}_{\sigma, n}}) - R(h^{*}) \leq \dfrac{ \mathcal{K}_{\ell}}{ n^{\frac{s}{s + d}(1-1/r) } }.
\end{equation}

\medskip

\noindent
\textbf{Step 3 : Expected excess risk bound}.
\medskip

\noindent
According to (\ref{Exp_excess_risk_thm2}),  (\ref{proof_exp_2_g_ing_thm2_v1}) and  (\ref{approxi_error2_thm2}), we have for all $n \geq n_0$,
\begin{align*}
\E[R(\widehat{h}_n) - R(h^{*})]  
& \leq \dfrac{C(\mu, \mathcal{K}_{\ell}, L_1, L_2)}{n^{\big(1 - 1/r \big)(\mu + 1)/(2\mu + 3)}} +  \dfrac{6\mathcal{K}_{\ell} M}{n^{(1- 1/r) (\mu+1)/(2\mu+3) }} + \dfrac{3\mathcal{K}_{\ell}}{n} + \dfrac{ 2\mathcal{K}_{\ell}}{ n^{\frac{s}{s + d}(1-1/r) } }.
\end{align*}

\qed

\end{document}